\pdfoutput=1
\documentclass[11pt]{article}

\usepackage[final]{acl}
\usepackage{times}
\usepackage{latexsym}
\usepackage[T1]{fontenc}
\usepackage[utf8]{inputenc}
\usepackage{microtype}
\usepackage{inconsolata}
\usepackage{graphicx}
\usepackage{url}
\usepackage{times}
\usepackage{latexsym}
\usepackage{longtable}
\usepackage{booktabs}
\usepackage{enumitem}
\usepackage{multirow}
\usepackage{tikz}
\usepackage{amsmath}
\usepackage[ruled, noend]{algorithm2e} 
\usepackage{tabularray}
\usepackage{centernot}
\usepackage{placeins}
\usepackage{amsfonts}
\usepackage[table]{colortbl}
\usepackage{tabularx}
\usepackage{lscape}
\usepackage{array}
\usepackage{cleveref}
\usepackage{tcolorbox}
\usepackage{hyperref}
\usepackage{xcolor}

\newcommand{\sref}[1]{\S\ref{#1}}
\newcommand{\method}{\textsc{ReCall}}
\title{\method: Membership Inference via Relative Conditional Log-Likelihoods}

\author{\\ \bf
    Roy Xie \hspace{4mm}
    Junlin Wang \hspace{4mm}
    Ruomin Huang \hspace{4mm}
    Minxing Zhang \hspace{4mm}
    Rong Ge \AND 
    \\ [-5.5ex] \bf 
    Jian Pei \hspace{4mm}
    Neil Zhenqiang Gong \hspace{4mm}
    Bhuwan Dhingra \\ [1ex]
    Duke University \\ [1ex]
\href{https://royxie.com/recall-project-page/}{royxie.com/recall-project-page}
}
\begin{document}
\maketitle

\begin{abstract}
The rapid scaling of large language models (LLMs) has raised concerns about the transparency and fair use of the data used in their pretraining. Detecting such content is challenging due to the scale of the data and limited exposure of each instance during training. We propose \method{}, (\textbf{Re}lative \textbf{C}ondition\textbf{a}l \textbf{L}og-\textbf{L}ikelihood), a novel membership inference attack (MIA) to detect LLMs' pretraining data by leveraging their conditional language modeling capabilities. \method{} examines the relative change in conditional log-likelihoods when prefixing target data points with non-member context. Our empirical findings show that conditioning member data on non-member prefixes induces a larger decrease in log-likelihood compared to non-member data. We conduct comprehensive experiments and show that \method{} achieves state-of-the-art performance on WikiMIA dataset, even with random and synthetic prefixes, and can be further improved using an ensemble approach. Moreover, we conduct an in-depth analysis of LLMs' behavior with different membership contexts, providing insights into how LLMs leverage membership information for effective inference at both the sequence and token level.
\end{abstract}

\section{Introduction}
The amount of pretraining data used to train large language models (LLMs) has quickly expanded in recent years, comprising of trillions of tokens sourced from a vast array of sources \citep{raffel2020exploring, brown2020language}. While such diversity and volume allow for a comprehensive language understanding, it also raises the concerns of including sensitive or unintended content such as copyrighted materials \citep{meeus2023did, duarte2024cop}, personally identifiable information \citep{tang2023privacy}, or test data from benchmarks \citep{oren2023proving, deng-etal-2024-investigating}. Additionally, the lack of transparency regarding the composition of pretraining datasets exacerbates these concerns, as many developers are reluctant to disclose full details due to proprietary reasons or the sheer volume of data involved.

To address these concerns, many works have proposed to detect pretraining data in LLMs \citep{mia_mink,mia_minkpp,mimir}, which involve using Membership Inference Attacks (MIAs) to infer whether a given data point was part of the training set. The basic MIA leverages a simple fact that member data are trained and memorized by the model, which leaves footprints in the model, resulting in higher log-likelihood (LL) than non-member data \citep{mia_loss}. However, the massive scale of pretraining data means that LLMs are typically trained for only a single epoch, making this problem particularly challenging \citep{mimir,mia_mink}, as each instance is only exposed once to the model, leading to limited memorization \citep{kandpal2022deduplicating, leino2020stolen}.

\begin{figure}[t]
% \centering
\includegraphics[width=0.485\textwidth]{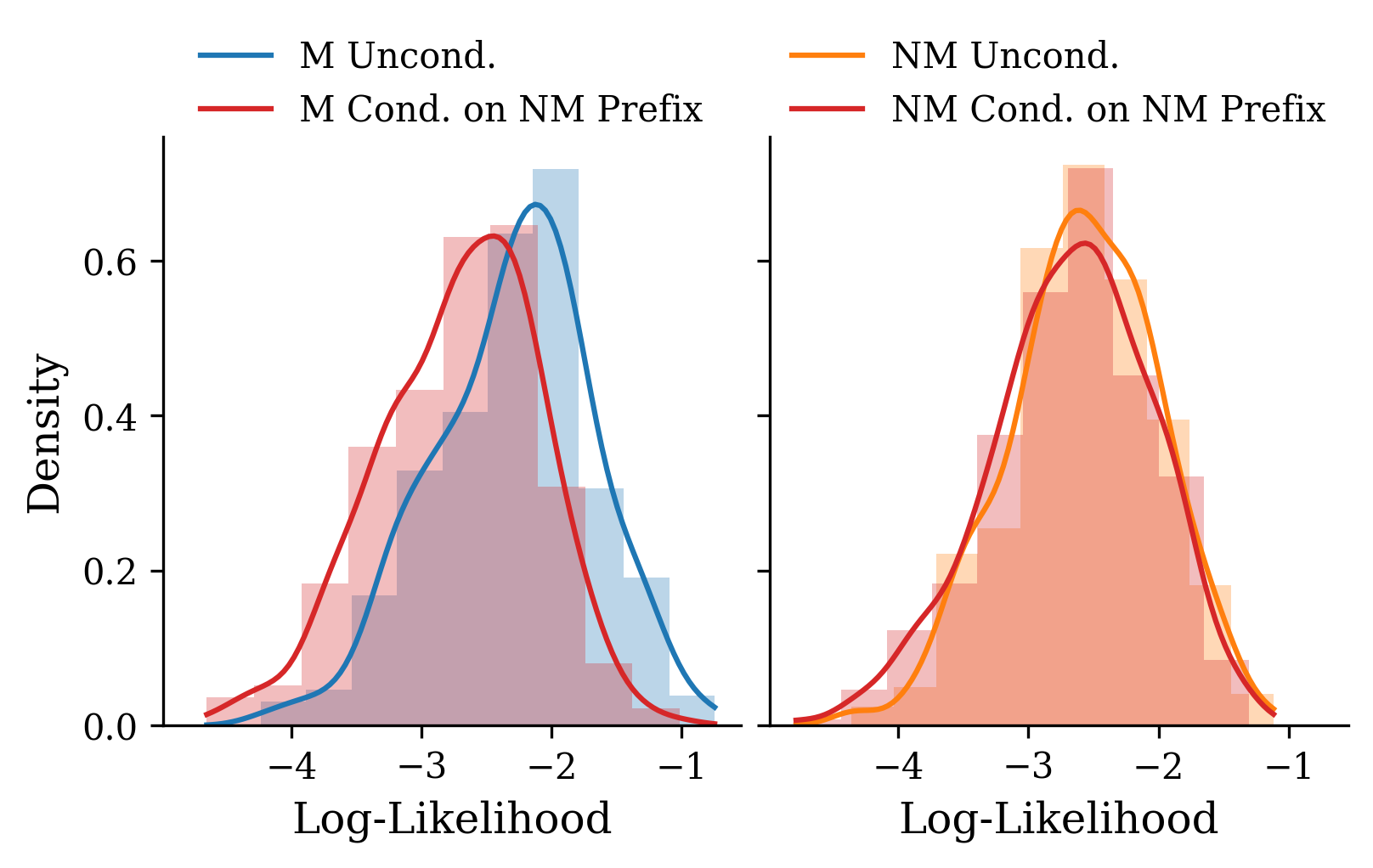}
\caption{Log-Likelihood comparison between members (M) and non-members (NM). Members experience a higher likelihood reduction than non-members when conditioned with non-member context.
}
\label{fig:cond_vs_uncond}
\vspace{-1em}
\end{figure}

In this work, we propose \method{} (\textbf{Re}lative \textbf{C}ondition\textbf{a}l \textbf{L}og \textbf{L}ikelihood), an efficient MIA detects LLMs pretraining data. \method{} leverages the conditional language modeling
capabilities of LLMs by examining the relative change in conditional LLs when prefixing data with non-member context. Our key empirical finding, as illustrated in Figure \ref{fig:cond_vs_uncond},\footnote{Plot result is from 5-shot setting using the Pythia-6.9B model on the WikiMIA-32 dataset. Additional visualizations are presented in \Cref{app:motivation_visual}.} is that conditioning member data on non-member prefix induces a larger decrease 
in LL compared to conditioning non-member data on other non-members. This observation forms the basis of \method{}. 
We leverage a few non-members from the target domain to construct the non-member prefix. While this might seem like a limitation, in practice, it is not a hard constraint. For most real-world applications, we can easily obtain non-member data points by selecting recent data that postdates the model's training data or by creating new, synthetic data points \citep{mia_mink, cheng2024dated,duarte2024cop}. 

One interpretation of this empirical finding comes from prior work on in-context learning (ICL), which suggests that it has an effect similar to fine-tuning \citep{akyurek2022learning}. By filling the context with non-members, we are essentially changing the predictive distribution of the language model. This change has a larger detrimental effect on members, which are already memorized by the model, compared to non-members, which the model is unfamiliar with regardless of the context. We further discuss this in \sref{sec:member_prefix}. 

We evaluate \method{} on two existing benchmarks, WikiMIA \citep{mia_mink} and MIMIR \citep{mimir}. WikiMIA provides different data length for fine-grained evaluation, while MIMIR presents a more challenging setting with minimal distribution shifts between members and non-members. Our comprehensive experiments demonstrate that \method{} achieves state-of-the-art performance on WikiMIA, outperforming existing MIA methods by a large margin, and obtains competitive results on MIMIR (\sref{sec:main_result}). We show that using random and synthetic prefixes achieves comparable performance to using real and optimal non-member data (\sref{sec:prefix_selection}). We propose an ensemble approach to further enhance the performance of \method{} and mitigate the limitations imposed by the fixed context window size of LLMs (\sref{sec:28_shots}). Lastly, we conduct an in-depth empirical investigation on how LLMs behave with different membership contexts at both the sequence and token level, providing insights into how LLMs leverage membership information for effective inference (\sref{sec:token_level}).

\section{Related Work}
\paragraph{Membership Inference Attacks}
Membership Inference Attacks (MIAs) aim to determine whether a given data sample was part of a model's training set. It was initially proposed by \citet{shokri2017membership} and has significant implication in tasks such as measuring memorization and privacy risk \citep{carlini2022quantifying, mireshghallah2022quantifying,steinke2024privacy}, serving as basis of advance attacks \citep{zlib,nasr2023scalable}, and detection on test-set contamination \citep{oren2023proving}, copyrighted content \citep{meeus2023did, duarte2024cop} and knowledge cutoff \citep{cheng2024dated} for LLMs. Research in MIA has been explored in natural language domain for both finetuning \citep{watson2021importance, mireshghallah2022quantifying, fu2023practical, mia_neighbor} and pretraining settings \citep{mia_mink, mimir, mia_minkpp}. Current LLMs typically train on the massive data for only a single epoch, which makes MIA more challenging compared to the multi-epoch finetuning setting \citep{carlini2022quantifying, mia_mink}. 

\paragraph{In-context Learning as Attack Vectors}
Transformer-based \citep{vaswani2017attention} LLMs, pretrained on vast amounts of data, have demonstrated a striking ability known as in-context learning (ICL) \citep{brown2020language}. Specifically, after pretraining, these models can learn and complete new tasks during inference without updating their parameters. In ICL, the model takes in a short sequence of supervised examples (prefix) from the task and then generates a prediction for a query example. Recently, ICL has been used as attack vectors for LLMs, such as jailbreaks \citep{wei2023jailbreak, anil2024many}, sensitive information extraction \citep{tang2023privacy}, and backdoor attacks \citep{kandpal2023backdoor}. In this work, we leverage a similar notion from ICL in an \textit{unsupervised} manner to conduct MIA by prefixing the target data points with non-member context. To the best of our knowledge, this is the first study to undertake such a task.

\section{\method{}: Relative Conditional Log-Likelihood}

\paragraph{Problem Definition}
Given that ${M}$ is an autoregressive language model that outputs a probability distribution over the next token given a prefix, let ${D}$ be a dataset used to train ${M}$. The goal of a membership inference attack is to determine, for a target data point $\mathbf{x}$, whether $\mathbf{x} \in {D}$ or $\mathbf{x} \notin {D}$. A membership score $S(\mathbf{x}; {M})$ is calculated and thresholded to classify whether $\mathbf{x}$ is a member or non-member of the training dataset ${D}$.

\begin{figure}[t]
\centering
\includegraphics[width=0.385\textwidth]{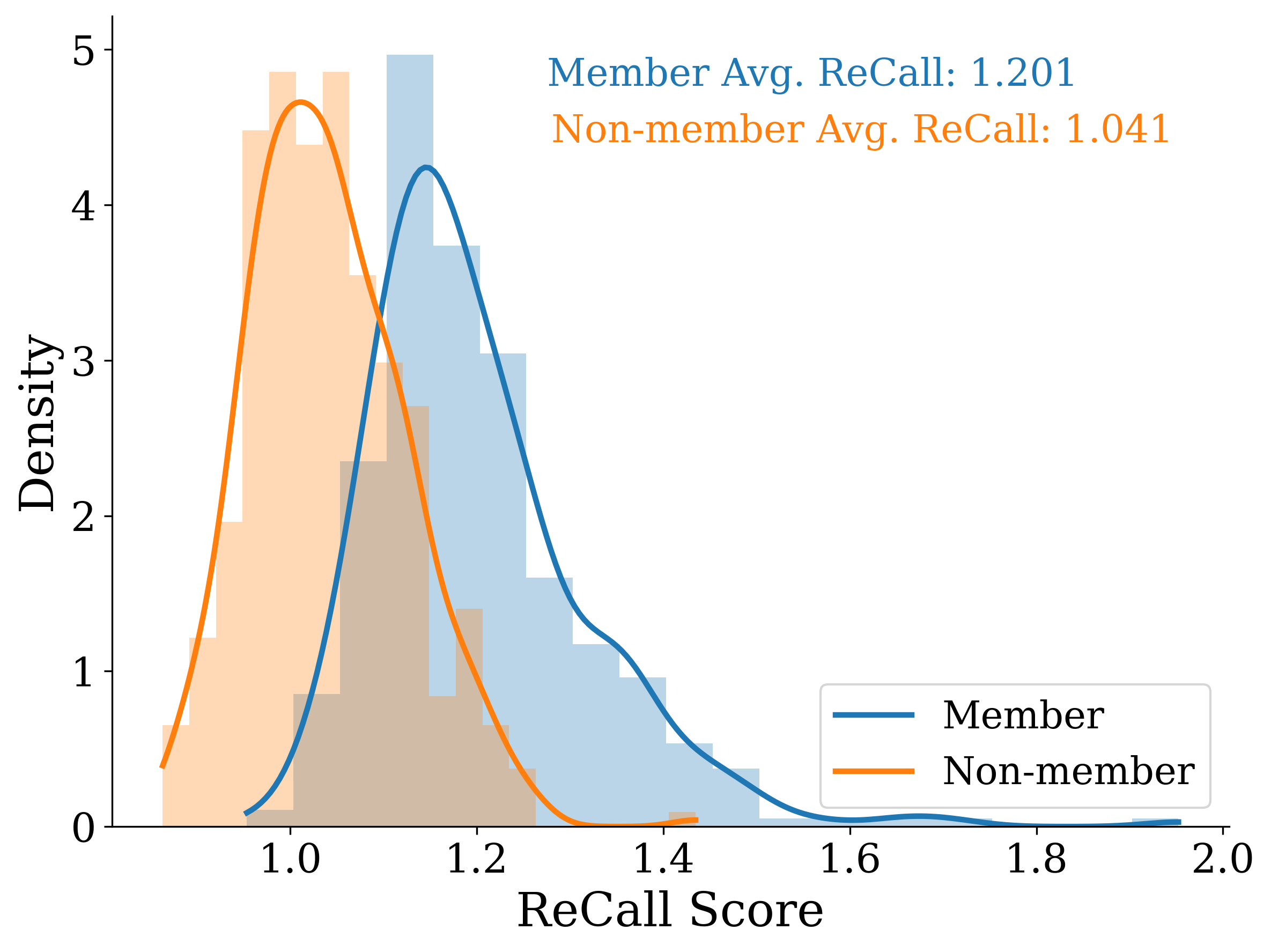}
\caption{Distribution of \method{} scores for members and non-members. Values close to 1 indicate changes are minimal. Overall, members tend to have higher \method{} scores compared to non-members. More visualizations can be found in \Cref{app:recall_visual_v1}. }
\label{fig:recall_violin}
\vspace{-1.5em}
\end{figure}

\paragraph{Proposed Method}
The key idea behind our proposed method is measuring the behavior of $M$ when conditioning the target data point with a non-member context (prefix). The \method{} score, which is the ratio of the conditional LL to the unconditional LL, is used to quantify this change. To begin, we select a prefix $P$, which is a sequence of non-member data points $p_i$ concatenated together:
\begin{equation}
P = p_{1} \oplus p_{2} \oplus ... \oplus p_{n}.
\end{equation}
The non-member data points are \textit{known} to be non-members of the model $M$. Non-member data can typically be obtained for LLMs based on knowledge cutoff time, or by using user-generated or machine-generated synthetic data \citep{mia_mink, cheng2024dated,duarte2024cop}, and we discuss the process of selecting prefixes in detail in \sref{sec:select_prefix}. For a given dataset $D$, the prefix $P$ is fixed. For each target data point $\mathbf{x}$, we calculate two LLs from $M$: (1) the unconditional LL of $\mathbf{x}$ itself, $LL(\mathbf{x})$, and (2) the LL of $\mathbf{x}$ conditioned on the prefix $P$, denoted as $LL(\mathbf{x}|P)$.\footnote{Note that LLs are \textit{negative} values. More information about LL can be found in \Cref{app:ll_explain}.} The \method{} score for target data point $\mathbf{x}$ is then calculated as:
\begin{equation}
    \method{} (\mathbf{x}) = \frac{LL(\mathbf{x}|P)}{LL(\mathbf{x})}.
\end{equation}
By providing $P$ to the model, we introduce additional unseen context and new knowledge without explicitly fine-tuning or updating the model parameters \citep{liu2023pre,brown2020language}. For member data points, denoted as $\mathbf{x_m}$, which have already been learned and memorized by the model, the introduction of unseen text may perturb the model's existing confidence to a larger scale compared to non-member data points, denoted as $\mathbf{x_{nm}}$. Consequently, as illustrated in \Cref{fig:recall_violin},\footnote{Plot result is from the same setting as \Cref{fig:cond_vs_uncond}.} we expect member data points to have \textit{higher} \method{} scores than non-member data points:
\begin{equation}\label{eqa:recall}
\mathbb{E}\left[\method{} (\mathbf{x_{\text{m}}})\right] > \mathbb{E}\left[\method{} (\mathbf{x_{\text{nm}}})\right].
\end{equation}
We provide a detailed discussion on the relationship between LL and \method{} scores in \Cref{app:ll_recall_explain}. 
As an inference-time algorithm, \method{} does not rely on access to the pretraining data distribution or a reference model, which are assumptions made by previous membership inference attacks \citep{first_principle,watson2021importance}. 
\paragraph{Prefix Selection}
The number of non-member data points $n$ used in $P$ (referred to as ``shots'') is the only hyperparameter in our method. The optimal number of $n$ may vary for different models since models have varying context window lengths \citep{jin2024llm}. Additionally, varying lengths of the target data points in $D$ can also affect the $n$ being used (\sref{sec:28_shots}). Generally, longer data points lead to better MIA performance, as they contain more information that can be memorized by the target model \citep{mia_mink}. We will demonstrate in \sref{sec:gpt4_prefix} that it is possible to use a random or synthetic prefix generated by LLMs to achieve high performance. In \sref{sec:28_shots}, we will show that only \textit{one} shot is needed for \method{} to outperform baselines.

\section{Experiments and Results}
In this section we conduct comprehensive experiments to demonstrate the effectiveness of \method{}. 

\subsection{Experimental Setup}
\paragraph{Benchmarks}
We focus on WikiMIA \citep{mia_mink} and MIMIR \citep{mimir} benchmarks. WikiMIA consists of text from Wikipedia, with member and non-member samples determined based on model's knowledge cutoff time. The dataset is grouped into splits based on sentence length (32, 64, 128) to enable fine-grained evaluation. MIMIR is derived from the Pile dataset \citep{pile} and covers various domains. To create a challenging setting for membership inference, MIMIR employs n-gram filtering to select member and non-member samples from the same dataset, maximizing their similarity \citep{mimir}. While this deviates from the standard MIA setting, we report results from both 13-gram and 7-gram MIMIR versions for a rigorous evaluation.

\paragraph{Baselines}
We compare \method{} against 6 state-of-the-art baselines: \textbf{\textit{Loss}} \citep{mia_loss} simplely uses input loss as the membership score; \textbf{\textit{Reference}} \citep{first_principle} calibrates input loss using a reference model; \textbf{\textit{Zlib}} \citep{zlib} compresses input loss using Zlib entropy; \textbf{\textit{Neighbor}} \citep{mia_neighbor} compares input loss to the average loss of similar tokens; \textbf{\textit{Min-K\%}} \citep{mia_mink} averages top-k\% minimum token probabilities from the input; and \textbf{\textit{Min-K\%++}} \citep{mia_minkpp} extends Min-K\% with normalization factors. More details can be found in \Cref{app:baseline_methods}.

\paragraph{Models}\label{models}
For WikiMIA, we experiment with a diverse set of transformer-based LLMs, including the Pythia 6.9B \citep{pythia}, GPT-NeoX 20B \citep{neox}, LLaMA 30B \citep{llama}, and OPT 66B \citep{opt}. We also include the Mamba model which uses a state space-based architecture \citep{mamba}. For MIMIR, we focus on the Pythia model family (160M, 1.4B, 2.8B, 6.9B, 12B parameters), consistent with \citep{mimir,mia_minkpp}. Following \citet{mia_mink} and \citet{mimir}, we use the smallest model version and the best performing reference model for Reference method. 

\paragraph{Metrics}
Following the standard evaluation procedure for MIAs \citep{mia_mink, mimir, mia_minkpp, mia_neighbor}, we use the area under the ROC curve (AUC) as our main evaluation metric, along with the true positive rate at a one percent false positive rate (TPR@1\%FPR) \citep{first_principle}. More details about the evaluation metrics can be found in \Cref{app:metric_detail}.

\paragraph{Implementation Details}\label{sec:implement_detail}
As all benchmarks do not provide validation set, and also following \citet{mia_minkpp}, we sweep over 1 to 12 for the number of shots and report the best result. We randomly select 12 data points as prefix candidates from the test set and exclude them from evaluation. We will show in \sref{sec:random_selection} that \method{} is robust to random selection. In \sref{sec:28_shots}, we will demonstrate that \method{} significantly outperforms all baselines even with just one shot. We also compare the \textit{best} possible performance of other methods for a fair comparison. More implementation details can be found in the \Cref{app:implement_detail}. 

\subsection{Main Results}\label{sec:main_result} 

\paragraph{WikiMIA Results}
\Cref{tab:wikimia_main_results} shows that \method{} achieves state-of-the-art performance on the WikiMIA benchmark. Our method consistently outperforms all existing baseline methods in all settings by a large margin. On average, \method{} surpasses the runner-up Min-K\%++ by {14.8\%, 15.4\%, 14.8\%} in terms of AUC scores for input lengths of {32, 64, 128}, respectively. Moreover, \method{}'s superior performance is consistent across different model architectures. The improvement is particularly significant for shorter inputs and smaller models, which are known to be more challenging for MIAs \citep{mia_mink}, which demonstrates the effectiveness of \method{} in capturing membership signals even in a challenging setting.
We also report the TPR@1\%FPR results in \Cref{app:wikimia_tpr}, which again shows the significant improvements and highlights the effectiveness of our approach in detecting pretraining data with high precision.

\begin{table*}[t!]
\centering
\resizebox{0.95\textwidth}{!}{%
\begin{tabular}{llcccccccc}
\toprule
\textbf{Len.} & \textbf{Method} & \textbf{Mamba-1.4B} & \textbf{Pythia-6.9B} & \textbf{LLaMA-13B} & \textbf{NeoX-20B} & \textbf{LLaMA-30B} & \textbf{OPT-66B} & \textbf{Average} \\
\midrule
\multirow{7}{*}{32}
& Loss & 60.7 & 63.6 & 67.6 & 68.7 & 69.5 & 65.4 & 65.9\\
& Ref & 60.9 & 63.7 & 67.7 & 68.9& 70.0  & 65.8 & 66.2\\
& Zlib & 61.6 & 64.1 & 67.8 & 68.9 & 69.9 & 65.5 & 66.3\\
& Neighbor & 64.1 & 65.8 &  65.8 & 70.2& 67.6 & 68.2& 66.9\\
& Min-K\% & 63.2 & 66.3 & 68.0 & 71.8& 70.1 & 67.4 & 67.8\\
& Min-K\%++ & 66.8 & 70.3 & 84.8 & 75.1 & 84.3 &  70.3& 75.3\\
\rowcolor{gray!15}& \method  &  \textbf{90.2} & \textbf{91.6} & \textbf{92.2} & \textbf{90.5}& \textbf{90.7} & \textbf{85.1} & \textbf{90.1}\\
\midrule
\multirow{7}{*}{64}
& Loss & 59.2 & 61.7 & 64.4 & 67.4  & 66.7 & 63.1 & 63.8\\
& Ref & 59.5 & 61.8 & 64.9 & 67.7 & 67.6 & 63.6 & 64.2\\
& Zlib & 61.4 & 63.3 & 65.9 & 68.7 & 68.0 & 64.6 & 65.3\\
& Neighbor & 60.6 & 63.2  & 64.1 & 67.1 & 67.1 & 64.1 & 64.4\\
& Min-K\% & 63.2 & 65.0 & 66.9 & 73.5 & 69.1 & 67.9 & 67.6\\
& Min-K\%++ & 67.2 & 71.7 & 85.6 & 76.0 & 84.8 & 70.2& 75.9\\
\rowcolor{gray!15}& \method  & \textbf{91.4} & \textbf{93.0} & \textbf{95.2} & \textbf{93.2} & \textbf{94.9} & \textbf{79.9} & \textbf{91.3} \\
\midrule
\multirow{7}{*}{128}
& Loss & 63.1 & 65.0 & 69.1 & 70.6 & 72.0 & 65.3 & 67.5\\
& Ref & 63.0 & 65.1 & 69.3 & 70.8 & 73.0 & 65.5 & 67.8\\
& Zlib & 65.5 & 67.8 & 71.5 & 72.6 & 73.6 & 67.6 &69.8\\
& Neighbor & 64.8 & 67.5 & 68.3 & 71.6 & 72.2 & 67.7 & 68.7\\
& Min-K\% & 66.8 & 69.5 & 71.5 & 75.0 & 74.2 & 70.2 & 71.2\\
& Min-K\%++ & 66.8 & 69.7 & 83.9 & 75.8 & 82.9 & 72.1 & 75.2\\
\rowcolor{gray!15}& \method  & \textbf{91.2} & \textbf{92.6} & \textbf{92.5} & \textbf{91.7} & \textbf{91.2} & \textbf{81.0} & \textbf{90.0}\\
\bottomrule
\end{tabular}
}
\caption{AUC results on WikiMIA benchmark. \textbf{Bolded} number shows the best result within each column for the given length. \method{} achieves significant improvements over all existing baseline methods in all settings.}
\label{tab:wikimia_main_results}
\end{table*}
\begin{table*}[ht!]
\begin{center} \scriptsize
\setlength{\tabcolsep}{0.7pt}
\begin{tabularx}{\textwidth}{l *{20}{>{\centering\arraybackslash}X}@{}}
    \toprule
    \multirow{2}{*}{}  & \multicolumn{5}{c}{\textbf{Wikipedia}} & \multicolumn{5}{c}{\textbf{Github}} & \multicolumn{5}{c}{\textbf{Pile CC}} & \multicolumn{5}{c}{\textbf{PubMed Central}} \\
    \cmidrule(lr){2-6}  \cmidrule(lr){7-11} \cmidrule(lr){12-16} \cmidrule(lr){17-21}
    \textbf{Method} & 160M & 1.4B & 2.8B & 6.9B & 12B
    & 160M & 1.4B & 2.8B & 6.9B & 12B
    & 160M & 1.4B & 2.8B & 6.9B & 12B
    & 160M & 1.4B & 2.8B & 6.9B & 12B
    \\
    \midrule
    
Loss & 50.2 & 51.3	& 51.8 & 52.8 &  53.5& 65.6& 69.5	& 71.0 &72.8  & 73.8 & 49.6 & 50.0	&50.1  &50.7  & 51.1 &50.0 &50.0	 & 50.2 & 50.9 & 51.5 \\
Ref & 50.9 & \textbf{54.7} & \textbf{57.6} & \textbf{60.3} & \textbf{61.7} & 63.9 & 67.0 & 65.3 & 64.3 & 63.1 & 48.8 & \textbf{52.3} & \textbf{53.7} & \textbf{54.6} & \textbf{56.4} & \textbf{51.0} & 52.1 & \textbf{53.6} & \textbf{55.9}& \textbf{58.1} \\
Zlib & 51.1 &  52.1&  52.5& 53.6 & 54.4& \textbf{67.4}& 	\textbf{70.8} & \textbf{72.1}  & 73.8 & 74.7 &49.5 & 	50.0&  50.2& 50.7 & 51.1 & 50.1& 	50.2 &  50.3&  50.9& 51.4 \\
Neighbor & 50.7 & 51.7 & 52.2 & 53.2 & / & 65.3 & 69.4 & 70.5 & 72.1 & / & 49.6 & 50.0 & 50.1 & 50.8 & / & 47.9 & 49.1 & 49.7 & 50.1 & / \\
Min-K\% & 49.8 & 51.3 & 51.5 & 53.2 & 54.3 & 64.4 & 68.8 & 70.3 & 72.2 & 73.4 & 50.2 & 51.0 & 50.5 & 51.3 & 51.4 & 50.4 & 49.9 & 50.5 & 51.0 & 52.4 \\
Min-K\%++ & 49.5 & 53.4 & 54.9 & 57.6 & 61.2 & 64.7 & 69.3 & 70.2 & 72.9 & 73.4 & 50.0 & 50.8 & 50.6 & 52.6 & 53.4 & 50.4 & 50.7 & 52.1 & 54.2 & 54.8 \\

\rowcolor{gray!15} \method &\textbf{51.3} & 52.3 & 52.3  & 54.0 &  54.6& 64.9 & 70.0& 71.7 & \textbf{74.2} & \textbf{74.8} & \textbf{50.9} &	51.7	& 50.2 &  51.6& 51.1 &49.9  &\textbf{52.3} &50.0 & 51.4& 53.4  \\
    
    \toprule
    \multirow{2}{*}{}  & \multicolumn{5}{c}{\textbf{ArXiv}} & \multicolumn{5}{c}{\textbf{DM Mathematics}} & \multicolumn{5}{c}{\textbf{HackerNews}} & \multicolumn{5}{c}{\textbf{Average}}\\
    \cmidrule(lr){2-6}  \cmidrule(lr){7-11} \cmidrule(lr){12-16} \cmidrule(lr){17-21}
    \textbf{Method} & 160M & 1.4B & 2.8B & 6.9B & 12B
    & 160M & 1.4B & 2.8B & 6.9B & 12B
    & 160M & 1.4B & 2.8B & 6.9B & 12B
    & 160M & 1.4B & 2.8B & 6.9B & 12B
    \\
    \midrule

Loss & 51.0 &  51.5	& 51.9 & 52.9 & 53.4 & 48.9 &48.5	& 48.4 &48.5  &48.5  & 49.4 &50.4 & 51.2& 51.9 & 52.6 &  52.1 &53.0 & 53.5 & 54.4 & 54.9\\
Zlib & 50.0 &  50.8	&  51.2& 52.2 & 52.6 & 48.2 &48.2&  48.1& 48.1 & 48.1 &  49.7&	50.2  & 50.7 &  51.1& 51.6 & 52.3 &53.2 &53.6 &54.3 &54.8\\
Ref & 50.0 & 51.6 & 53.5 & \textbf{56.0} & \textbf{57.8} & \textbf{51.4} & 51.4 & 50.7 & 51.6 & 51.3 & 49.5 & 52.3 & \textbf{55.6} & \textbf{57.9} & \textbf{60.9} & 52.2 & 54.5 & \textbf{55.7} & \textbf{57.2} & \textbf{58.5} \\
Neighbor & 50.7 & 51.4 & 51.8 & 52.2 & / & 49.0 & 47.0 & 46.8 & 46.6 & / & {50.9} & {51.7} & 51.5 & 51.9 & / & 52.0 & 52.9 & 53.2 & 53.8 & / \\
Min-K\%  & 51.7 & 52.0 & 53.1 & 53.7 & 55.2 & 50.3 & 50.0 & 50.0 & 49.7 & 50.2 & 50.9 & 51.9 & 52.4 & 53.6 & 54.7 & 52.5 & 53.6 & 54.0 & 55.0 & 55.9 \\
Min-K\%++ & 50.7 & 51.0 & \textbf{53.9} & 55.5 & 58.4 & 50.9 & 49.8 & \textbf{ 51.8} & \textbf{52.0} & \textbf{52.1} & 50.7 & 51.2 & 52.9 & 54.6 & 56.8 & 52.4 & 53.7 & 55.3 & 57.0 & \textbf{58.5} \\

\rowcolor{gray!15} \method  &\textbf{52.5} &\textbf{52.8}& 52.7 &  54.6& 55.9 &50.9  & \textbf{52.8}& {51.3} &50.8  &50.8  & \textbf{52.4} &	\textbf{53.0} & {53.2}  & 54.2 &  54.7& \textbf{53.3} & \textbf{54.6}& 54.5 & 55.8 & 56.5\\
    
\bottomrule
\end{tabularx}
\caption{AUC results on MIMIR benchmark. \method{} achieves competitive performance compared to state-of-the-art methods, especially on smaller models (160M and 1.4B), while not relying on any reference models. The best results for each dataset and model size are highlighted in \textbf{bold}.}
\label{tab:mimir_main_results_13}
\vspace{-1.5em}
\end{center}
\end{table*}

\paragraph{MIMIR Results}
On the more challenging MIMIR benchmark, \method{} achieves competitive performance compared to state-of-the-art methods, as shown in \Cref{tab:mimir_main_results_13} (13-gram). It is important to note that MIMIR presents a much more challenging scenario as it deviates from the standard membership inference setting by minimizing the distribution shift between members and non-members \citep{maini2024llm,mimir}. The AUC scores for the 13-gram setting for all MIAs are close to \textit{random guessing}, indicating the difficulty for MIAs when members and non-members are very similar \citep{mimir}. Despite this, \method{} on average outperforms all baselines on 160M and 1.4B models. For other models, the Reference method dominates, but it is important to note that exhaustively searching for the best reference model among different candidate models is not only computationally expensive but also may not be feasible in practice \citep{mimir}. In contrast, \method{} does not rely on any reference models, yet still provides competitive performance. The 7-gram results in \Cref{app:mimir_main_results_7} show better performance than the 13-gram setting, with \method{} achieving the highest AUC  on 1.4B, 2.8B, 6.9B, and 12B models. We also report the TPR@1\%FPR results for both settings in \Cref{app:mimir_tpr_13}.

\section{Analysis}\label{sec:select_prefix}
In this section, we conduct a series of investigations to better understand \method{}. Following \citet{mia_minkpp}, we focus on the WikiMIA using the Pythia-12B model for our analysis. 

\subsection{Prefix Selection}\label{sec:prefix_selection}

\paragraph{Dynamic Prefix with Different Similarities} 
We investigate if using prefixes that are similar to the target data points results in better performance. For each data point, we search the entire dataset and create prefixes based on the Term Frequency-Inverse Document Frequency (TF-IDF) similarity scores \citep{sparck1972statistical}: (i) most similar (highest scores), (ii) moderately similar (middle scores), (iii) least similar (lowest scores), and (iv) random selection (random scores). In this setting, each target data point has its own prefix.

We compare the results with the original fixed-prefix setting\footnote{Using one fixed prefix for all target data points.} in \Cref{tab:dynamic_prefix}. The results indicate that using the most similar prefix yields the best performance, followed by random selection, moderate similarity, and least similar prefix. 
This suggests that the most effective prefixes are those similar to the target data point, with random selection providing the next best performance. We also observe that dynamic prefix selection does not perform as well as using a fixed prefix, likely because each data point creates a different threshold when using a dynamic prefix, leading to inconsistencies.

\begin{table}[t!]
\centering
\small
\begin{tabular}{lccc}
\toprule
\textbf{Similarity} & \textbf{Len. 32} & \textbf{Len. 64} & \textbf{Len. 128} \\
\midrule
Random  & 69.0 & 71.9 & 74.2 \\
Least  & 57.2 & 69.6 & 61.0 \\
Moderate  & 66.9 & 71.6 & 70.2 \\
Most  & 74.1 & 76.1 & 77.6 \\
\midrule
Fixed & 88.2 & 88.8 & 87.8 \\
\bottomrule
\end{tabular}
\caption{\method{} perform better with fixed prefix than dynamic prefix. Similar prefix results best performance, followed by random selection.}
\label{tab:dynamic_prefix}
\vspace{-1em}
\end{table}

\paragraph{Randomly Selected Prefix}\label{sec:random_selection}
We investigate the impact of randomly selecting non-member prefixes on the performance of \method{}. We randomly select 12 non-member data points from the test set, divide them into 3 sets, and compare the results with the best-performing baselines, Min-K\% and Min-K++\%, in \Cref{tab:random_selection}. The results show that the performance of \method{} across all groups is similar, with an average difference of 2.5\% between the top and lowest prefix, while significantly outperforming the baselines. This finding suggests that \method{} is robust to random prefix selection, as long as the prefixes are indeed non-members. In \sref{sec:member_prefix}, we will show that while their similarity to the target data point is preferred, the effectiveness of \method{} appears to be largely dependent on the non-member status of the prefixes. 

\begin{table}[ht]
\small
\centering
\begin{tabular}{lccc}
\toprule
\textbf{Prefix Set} & \textbf{Len. 32} & \textbf{Len. 64} & \textbf{Len. 128} \\
\midrule
Set 1 & 88.2 & 88.8 & 87.8 \\
Set 2 & 90.4 & 91.4 & 90.5 \\
Set 3 & 87.7 & 89.4 & 89.2 \\
\midrule
Min-K\% & 67.7 & 67.9 & 70.2 \\
Min-K\%++  & 72.4 & 72.5 & 72.7 \\
% Original & 88.2 & 88.8 & 87.8 \\
\bottomrule
\end{tabular}
\caption{AUC scores of \method{} with randomly selected prefixes divided into three sets, compared to the best-performing baselines, Min-K\% and Min-K++.}
\label{tab:random_selection}
\vspace{-1em}
\end{table}

\begin{table}[ht]
\small
\centering
\begin{tabular}{lccc}
\toprule
\textbf{Prefix Domain} & \textbf{Len. 32} & \textbf{Len. 64} & \textbf{Len. 128} \\
\midrule
GitHub & 72.4 & 69.5 & 68.8 \\
arXiv & 73.4 & 72.0 & 72.0 \\
Wikipedia & 83.3 & 87.1 & 86.5 \\
Original & 88.2 & 88.8 & 87.8 \\
\midrule
Min-K\% & 67.7 & 67.9 & 70.2 \\
Min-K\%++  & 72.4 & 72.5 & 72.7 \\
\bottomrule
\end{tabular}
\caption{AUC scores of \method{} when using prefixes from different domains obtained based on model knowledge cutoff time. Similar domains are preferred. 
}
\label{tab:out_of_domain_prefix} \vspace{-1em}
\end{table}

\paragraph{Varying Domain Prefix}
We explore the impact of prefix similarity on \method{}'s at a domain-level. The WikiMIA dataset was constructed using knowledge cutoff, with Wikipedia articles published after the model's training data used as non-members. We randomly select prefixes from Wikipedia (most similar), arXiv (moderately similar), and GitHub (least similar) and present the results in \Cref{tab:out_of_domain_prefix}.
We observe that when the prefix domain is significantly different from the target data (GitHub), \method{}'s effectiveness is reduced. However, the Wikipedia prefix achieves performance close to the original dataset, indicating that \method{} is not overfitting to the peculiarity of the original dataset. This suggests that selecting prefixes from the same or similar domains as the target data is preferred.
While the original WikiMIA dataset and the Wikipedia prefix setting both use knowledge cutoff, there might be differences in the specific passages selected. However, the strong performance of the Wikipedia prefix demonstrates that \method{} can generalize well to other non-member data points from the same domain, providing further evidence of its effectiveness.

\paragraph{GPT-4o Generated Prefix}\label{sec:gpt4_prefix}
The assumption of accessing a small number of non-member data points as prefix might not always be feasible as the membership of the selected prefix \textit{itself} could be mixed or unknown in a real-world scenario. Therefore, we explore the possibility using synthetic prefixes generated by LLMs from a mix of members and non-members. We randomly select 6 members and nonmembers and generate a synthetic prefix using GPT-4o \citep{GPT-4o} based on them.\footnote{Prompt can be found in \Cref{app:gpt_prompt}.} We compare results to the baselines and using ground-truth non-member data point as prefix in \Cref{tab:gpt_prefix}. We observe that even with the synthetic prefix from a mix of members and nonmembers, the performance is still close to the original setting, where the ground-truth non-members are used. This highlights the potential of using synthetic prefixes in situations where access to ground-truth non-member data is limited or unavailable, expanding the applicability of \method{} to a practical scenario. More synthetic prefixes results can be found in \Cref{app:gpt_prefix}.

\begin{table}[ht!]
\small
\centering
\begin{tabular}{lccc}
\toprule
\textbf{Prefix Setting} & \textbf{Len. 32} & \textbf{Len. 64} & \textbf{Len. 128} \\
\midrule
Synthetic & 85.4 & 90.3 & 86.4 \\
Original & 88.2 & 88.8 & 87.8 \\
\midrule
Min-K\% & 67.7 & 67.9 & 70.2 \\
Min-K\%++  & 72.4 & 72.5 & 72.7 \\
\bottomrule
\end{tabular}
\caption{AUC scores of \method{} with synthetic prefixes generated from GPT-4o, compared to the prefix from original dataset and two best-performing baselines. }
\
\label{tab:gpt_prefix}
\vspace{-1em}
\end{table}

\begin{figure*}[t!]
    \centering
    \includegraphics[width=\textwidth]{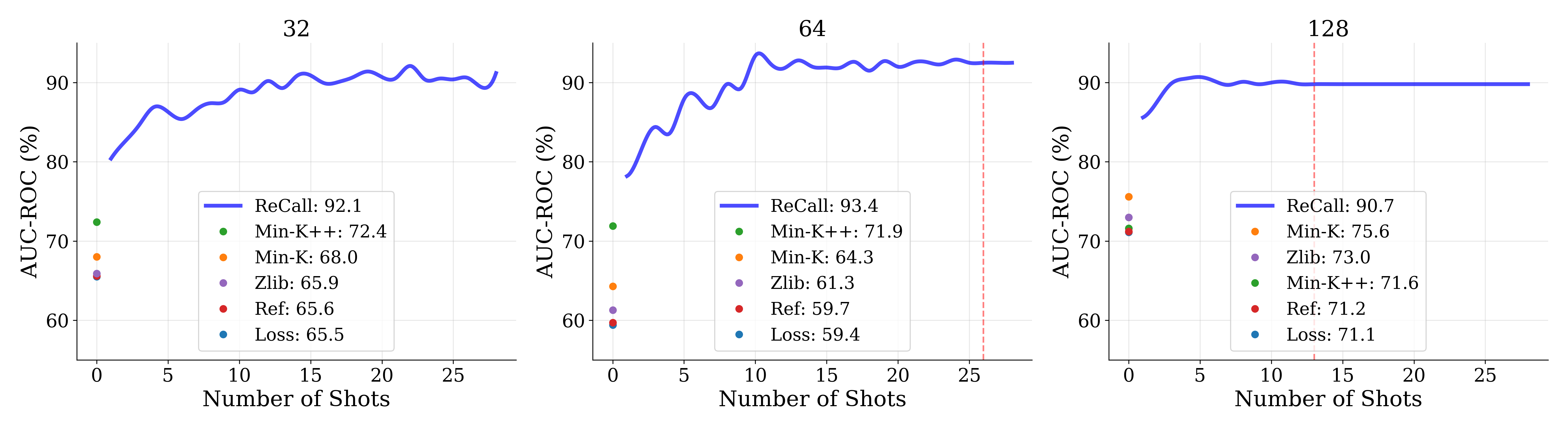}
\caption{\method{} performance up to 28 shots. \textcolor{red}{Red} dash line represents the LLMs' context window limit. \method{} consistently outperforms baselines across all settings, even with just \textit{one} shot.}
\label{fig:28_shots}
\end{figure*}

\subsection{Shots vs. Performance}\label{sec:28_shots}

\paragraph{Impact of Number of Shots and Context Window Size}
In general, increasing the context length improves the performance of MIAs, as the model can leverage more information to distinguish between members and non-members \citep{mia_mink}. However, LLMs have fixed context window sizes that limit the amount of text they can process in a single input. 
Exceeding the context length resultsing in remaining at the maximum number of shots which can fit in the context, so the performance should plateau after a while. We evaluate the performance of \method{} with up to 28 shots, intentionally exceeding the context window to probe the limitation. We compare the results to baselines in \Cref{fig:28_shots} and observe that \method{} consistently outperforms all baselines by a significant margin, even with just \textit{one} shot. As the number of shots increases, \method{}'s performance improves across all settings. As expected, the performance plateau when the length of the input exceeds the context window limit, which can be observed in the 64 and 128 length settings. This is because the longer the data length, the fewer shots are needed to exceed the context window.

\begin{table}[ht!]
\small
\centering
\begin{tabular}{lccc}
\toprule
\textbf{Prefix Setting} & \textbf{Len. 32} & \textbf{Len. 64} & \textbf{Len. 128} \\
\midrule
28-shot & 92.1 & 93.4 & 90.7 \\
Ensemble & 93.0  & 94.9  & 91.5 \\
\midrule
Min-K\% & 67.7 & 67.9 & 70.2 \\
Min-K\%++  & 72.4 & 72.5 & 72.7 \\
\bottomrule
\end{tabular}
\caption{
Performance comparison of the ensemble method, 28-shot method, and baselines. By taking the average of the \method{} scores using ensemble method, we can make a more robust prediction. 
}
\label{tab:ensemble_comparison}
\vspace{-1em}
\end{table}

\paragraph{Further Improvement with Ensemble Method} 
When dealing with a large number of shots, the context window limit of LLMs can be a bottleneck for the performance of \method{}. To circumvent this problem, we propose an ensemble method. Instead of using all 28 shots at once, we divide them into smaller sets as prefixes and calculate the \method{} score for the target data point under each set. We then take the mean of these independent \method{} scores to obtain the final score. The intuition behind this approach is that each group provides an independent \method{} score on the membership status of the input text, and by averaging them, we can reduce the variance and obtain a more robust estimate. We compare the performance of the ensemble method with the 28-shot method and the baselines in \Cref{tab:ensemble_comparison}. The results show that the ensemble method provides further improvement over the 28-shot method for all settings, demonstrating \method{}'s utility in leveraging a large number of shots while respecting the context window limit.

\begin{figure}[ht!]
\centering
\includegraphics[width=0.45\textwidth]{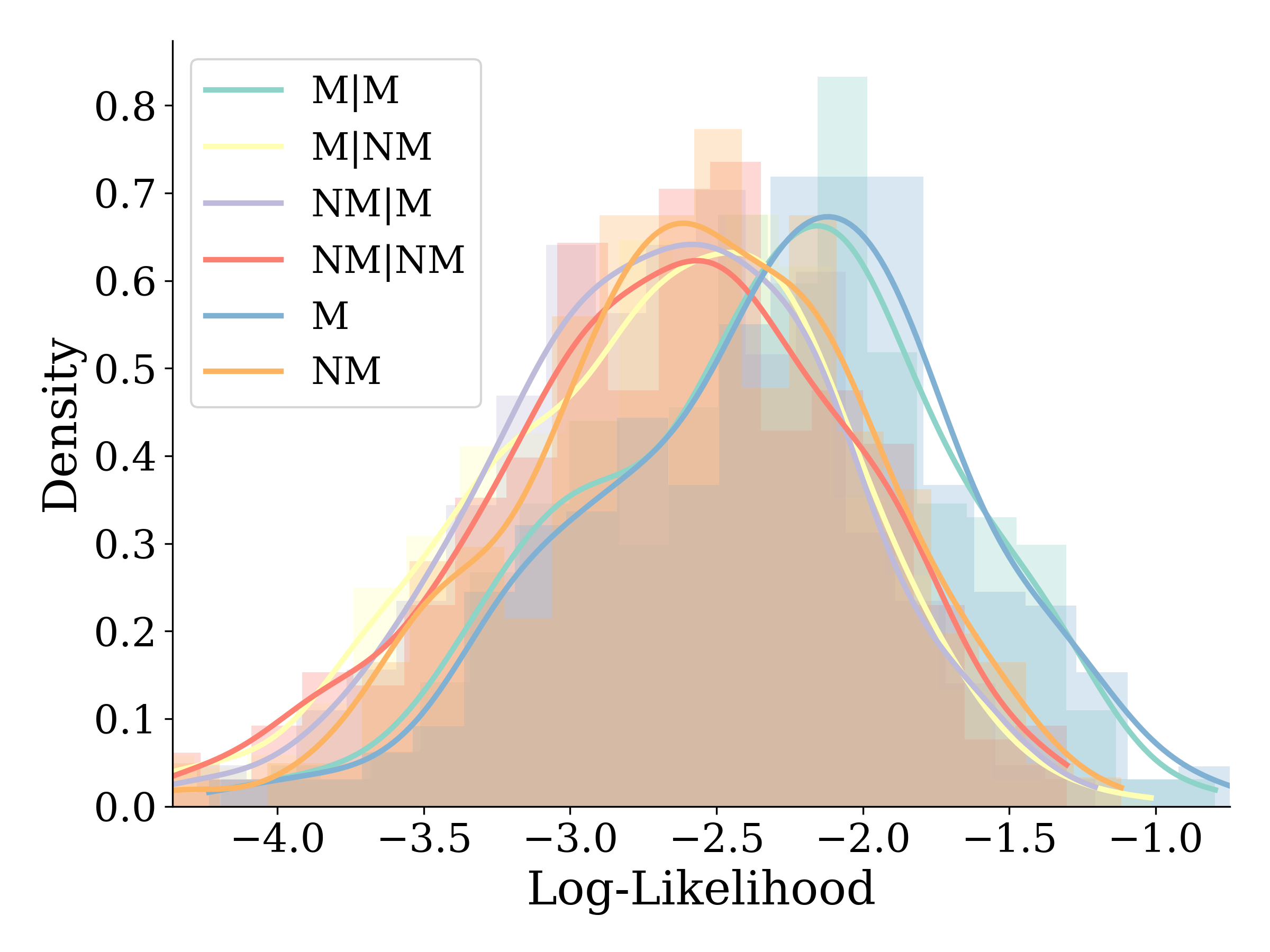}
\caption{
Conditioning both member and non-member with member prefix do not yield significant changes in LL compare to non-member prefix. More visualization can be found in \Cref{app:member_prefix_visual}.
}
\label{fig:member_prefix}
\vspace{-1em}
\end{figure}

\subsection{Discussions}
\paragraph{Why Non-member Prefixes?} \label{sec:member_prefix}
While non-member data can be easily accessed based on the data characteristics such as knowledge cutoff time, access to member data is much harder to assume, as LLMs training data are usually not disclosed \citep{mia_mink}. We empirically show that using member prefix is not only an unrealistic assumption but also does not yield the desired effect for detecting pretraining data. Following the same setting as \Cref{fig:cond_vs_uncond}, we prefix the target data points with both member and non-members and present the results in \Cref{fig:member_prefix}. We observe that conditioning both member and non-member data with a member prefix does not result in significant changes in LL compared to their unconditional LL. This suggests that using member data as context does not induce the distribution shift necessary for \method{} to effectively distinguish between members and non-members. We hypothesize that prefixing with additional member data does not significantly alter the model's predictive distribution because the model has already memorized the member data during pretraining and is familiar with its distribution. In contrast, prefixing with non-member data introduces a distribution shift that has a more pronounced effect on the LL of member data compared to non-member data. These findings demonstrate that \method{} is indeed leveraging the membership information of the prefix data to make prediction.

\begin{figure}[t]
\centering
\includegraphics[width=0.44\textwidth]{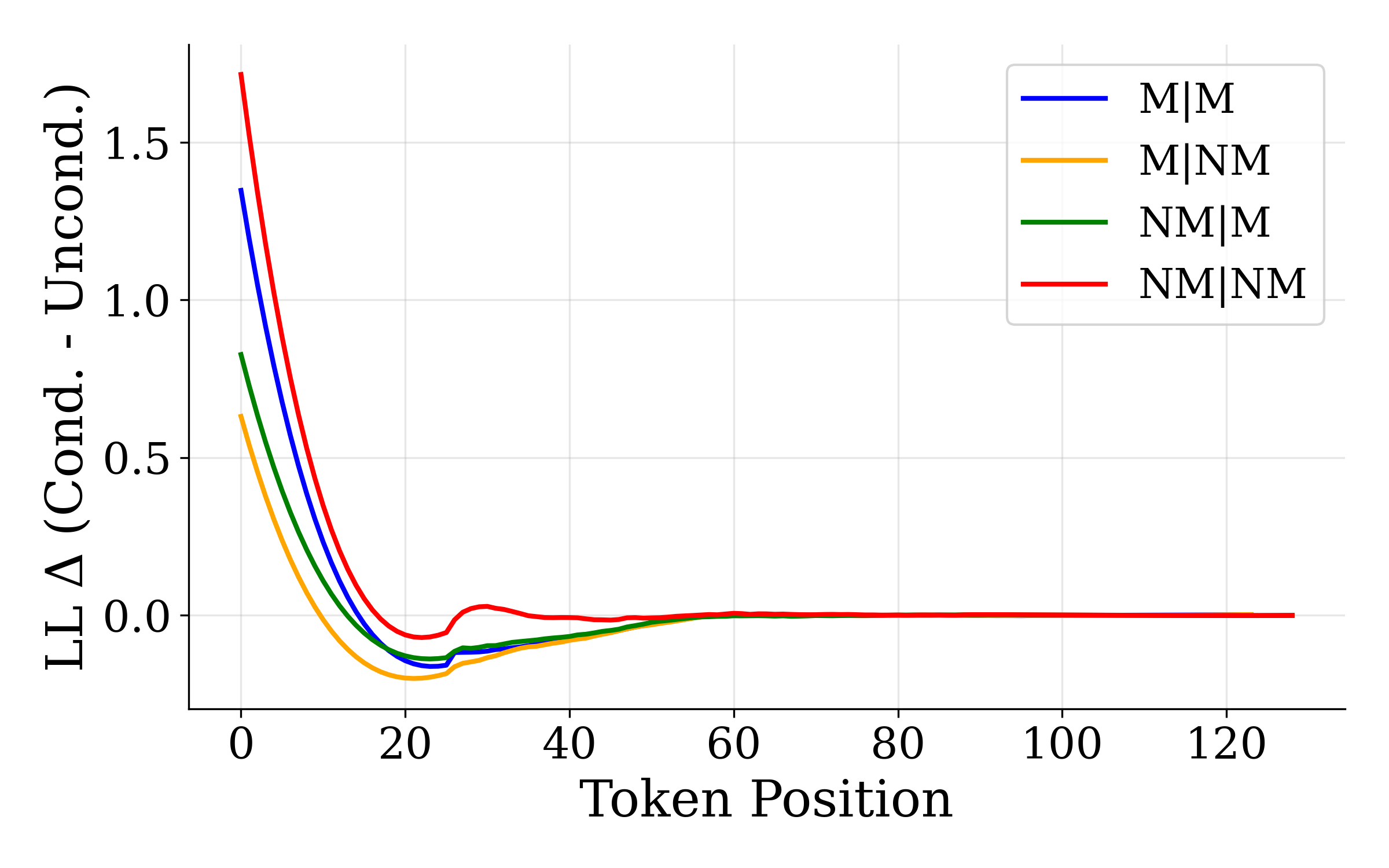}
\caption{Token-level LL changes for members and non-members with different membership prefix. The largest changes occur in the beginning tokens. 
Member and non-member data are most different when prefixed with non-member context. }
\label{fig:token_levelL_visual}
\vspace{-1em}
\end{figure}

\paragraph{Token-level Analysis}\label{sec:token_level}
Previous works, such as \citet{mia_mink} and \citet{mia_minkpp}, have leveraged token-level signals for MIAs. Similarly, we investigate \method{} at the token level with both member and non-member prefixes. We examine where the changes are occurring and how member and non-member prefixes impact token-level LL. For each token position, we take the average from all data points and present the LL change in \Cref{fig:token_levelL_visual} and more in \Cref{app:token_level_visual}. We observe three interesting points: (i) Most changes occur in the beginning tokens for all settings, especially the first few tokens. This is because the model becomes more confident about predicting the next token as it approaches the end of the sequence, given that it has already seen the preceding context. (ii) The changes are most dominant when a data point is prefixed with context from the same membership (e.g., M|M and NM|NM). This means the model has a stronger preference to continue with text from the same membership status. (iii) The differences between NM|NM and M|NM are more pronounced than those between M|M and NM|M, which further supports our finding that using non-member prefixes is effective for \method{} to distinguish between members and non-members, while member prefixes do not yield desired performance.

\paragraph{MIA Evaluation}
Recently, the MIA community has been discussing the effectiveness of MIA for LLMs \citep{mimir,blind}. Two key challenges complicate MIA evaluations: the vast scale of pretraining data and the temporal distribution shift between members and non-members. While \citet{blind} demonstrates that simple text classifiers trained \textit{directly} on the dataset can achieve superior performance, the equivalence of using model internal logits versus direct text classification remains an open question.
We conduct additional experiments on two applicable datasets used by \citet{blind}: Temporal Wiki and Temporal arXiv \citep{mimir}. We present results in Table \ref{tab:blind_compare} and observe that \method{} surpasses text classifier performance on both Temporal Wiki and Temporal arXiv datasets, while the text classifier shows superior results on WikiMIA. These results indicate that \method{} effectively discriminates between members and non-members, particularly for datasets structured based on temporal distributions.

\begin{table}[h!]
\small
\centering
\begin{tabular}{lcc}
\toprule
\textbf{Dataset} & \textbf{Text Classifier} & \textbf{\method{}} \\
\midrule
WikiMIA  & 98.7 & 95.2 \\
Temporal Wiki & 79.9 & 81.2 \\
Temporal arXiv & 75.6 & 76.0 \\
\bottomrule
\end{tabular}
\caption{AUC score comparison between text classifier and \method{} for three different datasets.}
\label{tab:blind_compare}
\vspace{-1em}
\end{table}

\section{Conclusion}
We introduced \method{}, a novel MIA for detecting pretraining data in LLMs by leveraging their conditional language modeling capabilities. \method{} captures the relative change in conditional log-likelihoods when prefixing target data points with non-member context. Through extensive experiments on WikiMIA and MIMIR benchmarks, we demonstrated \method{}'s state-of-the-art performance, outperforming existing MIA methods on WikiMIA. We showed that random and synthetic prefixes achieve comparable performance to real non-member data, enhancing \method{}'s practicality. \method{} consistently outperforms baselines and can be further improved using an ensemble method. Our in-depth analysis revealed valuable insights into LLMs' behavior under different membership contexts. As future work, we plan to investigate the theoretical aspects of \method{} and explore more efficient MIA methods.

\section{Limitations}
While \method{} demonstrates strong empirical performance, the theoretical analysis of why it works is limited in this work. We provide some hypotheses based on the connections between ICL and conditional language modeling in LLM, but a more rigorous and in-depth theoretical investigation is needed to fully understand the underlying mechanisms. This is particularly important given that ICL itself is an \textit{understudied} area, and the research community is still actively exploring how and why it works \citep{wei2023larger,liu2023pre,anil2024many}. A better understanding of ICL could provide valuable insights into our method. We believe that further theoretical analysis of \method{} and its interplay with ICL is an important direction for future research.
Our method assumes gray-box access to the target model, which requires access to its output probabilities. However, it is important to note that this limitation is shared by existing pretraining data detection methods \citep{mia_mink,mia_minkpp,mimir,mia_neighbor,first_principle, mia_loss}. In the future, we plan to explore methods that require less access to the target model.

\section{Ethics Statement}
Our primary intention is to advance the detection of sensitive content in LLMs, which is important for protecting privacy and intellectual property. However, we acknowledge that, like any tool, it could be misused to extract private information. Protecting user privacy should be a key priority as LLMs become increasingly ubiquitous and powerful. We call for further research on privacy-preserving LLM development and strategies to prevent misuse. 

\section{Acknowledgment}
We thank Jack Parker and Jabari Kwesi for helpful discussions. We also thank anonymous reviewers for their insightful feedback. This work was supported in part by the NSF Graduate Research Fellowship and NSF grants No. 2112562, 1937787, 2131859, 2125977, 2331065. 

\bibliography{custom}

\clearpage
\newpage
\appendix
\onecolumn

\section{Conditional and Unconditional LL Visualizations}\label{app:motivation_visual}
\begin{figure}[ht!]
\centering
\includegraphics[width=\textwidth]{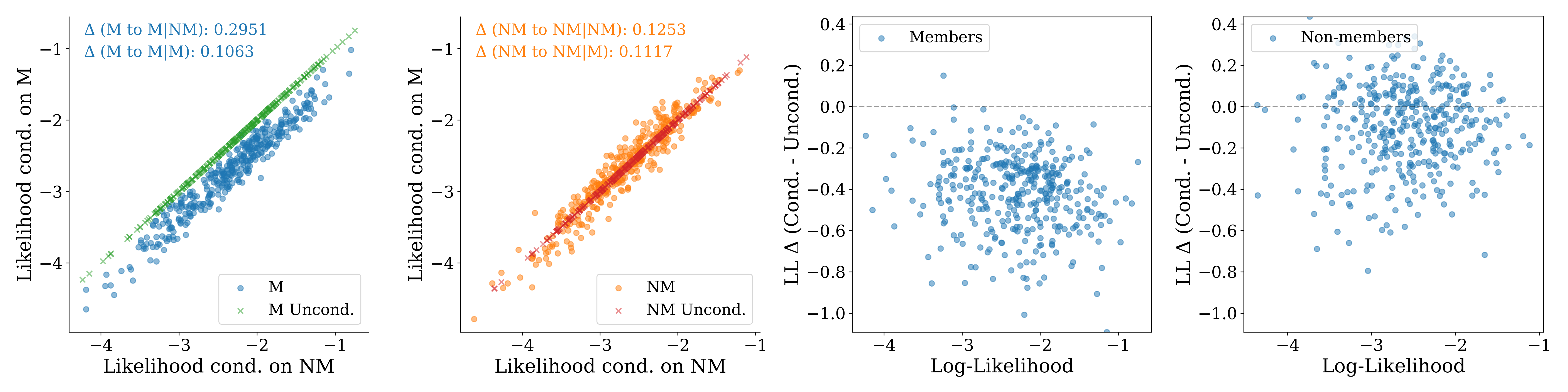}
\caption{Log-likelihood comparison with 5-shot non-member prefix for Pythia-6.9B model on WIKIMIA dataset. \textbf{Left Two}: shows distribution the conditional and unconditional version of both members and non-members. \textbf{Right Two}: shows the LL difference between the conditioned and unconditioned version of the data. Member data is experiencing much higher distributional shift than non-member data. }
\end{figure}
\section{Additional Log-Likelihoods and \method{} Scores Details}

\subsection{Log-Likelihood}\label{app:ll_explain}
Log-likelihood (LL) can be used to measure of how likely a given text is trained under a specific language model. A higher LL indicates that the model is more confident in predicting the text, suggesting that the model might have been trained such data. Conversely, a lower LL implies that the model is less familiar with the text. LL is closely related to other metrics such as loss and perplexity, where a higher LL corresponds to a lower loss and perplexity. While metrics like perplexity or probability can be used for membership inference, we use LL in this work since it is more numerically stable and mitigates underflow problems \citep{goodfellow2016deep}.

\subsection{Relationship between LL and \method{} Scores}\label{app:ll_recall_explain}
We show that conditioning member data on non-member prefixes induces a larger decrease in LL compared to non-member data, as suggested by Figure 1. We introduce \method{} score as our membership score to quantify this change. For a member data point $x_m$, the \method{} score is typically greater than 1, as the conditional LL is generally lower than the unconditional LL:

\begin{equation}
\method{}(x_m) = \frac{LL(x_m|P)}{LL(x_m)} > 1, \text{ since } LL(x_m|P) < LL(x_m) < 0.
\end{equation}

Note that LLs are \textbf{negative} values. For a non-member data point $x_{nm}$, the \method{} score can be either greater than, equal to, or less than 1. In some cases, prefixing a non-member data point with a non-member prefix might increase the model's confidence, resulting in a \method{} score less than 1. However, the core idea of \method{}, as illustrated in \Cref{fig:recall_violin} and \Cref{app:member_prefix_visual}, is that for member data points, the \method{} score is \textit{consistently higher} than non-member data points regardless the LL changes (increase or decrease) in non-members (\Cref{eqa:recall}). To better illustrate this, consider a member data point $x_m$ with $LL(x_m|P) = -4$ and $LL(x_m) = -3$. The \method{} score for $x_m$ is calculated as:

\begin{equation}
\method{}(x_m) = \frac{LL(x_m|P)}{LL(x_m)} = \frac{-4}{-3} = 1.3
\end{equation}

Note that $LL(x_m|P) < LL(x_m)$, indicating a decrease in LL when conditioning on the non-member prefix $P$. Now, consider a non-member data point $x_{nm}$ with $LL(x_{nm}|P) = -3.3$ and $LL(x_{nm}) = -3$. The \method{} score for $x_{nm}$ is:

\begin{equation}
\method{}(x_{nm}) = \frac{LL(x_{nm}|P)}{LL(x_{nm})} = \frac{-3.3}{-3} = 1.1
\end{equation}

% Again, $LL(x_{nm}|P) < LL(x_{nm})$, showing a decrease in LL when conditioning on $P$.

Comparing the member and non-member data points, we observe that $LL(x_m|P) < LL(x_{nm}|P)$, indicating a larger decrease in LL for the member data point when conditioned on the non-member prefix. However, the \method{} score for the member data point is higher than that of the non-member data point: $\method{}(x_m) > \method{}(x_{nm})$.
\section{Additional Baseline Details}\label{app:baseline_methods}
Given target data point $\mathbf{x}$, a MIA aims to determine if $\mathbf{x}$ was part of the training dataset ${D}$ used to train a model ${M}$ by computing a membership score $S(\mathbf{x}; {M})$. We provide a detailed description of baseline MIA methods used in our experiments. For each method, we explain how the membership score is calculated and the intuition behind the approach. 

\subsection{LOSS}
The LOSS baseline \citep{mia_loss} uses the model's computed loss over the target sample as the membership score. The intuition behind this approach is that the model will have lower loss values for data points it has seen during training (members) compared to unseen data points (non-members).

\begin{equation}
S(\mathbf{x}; {M}) = {Loss}(\mathbf{x};{M})
\end{equation}

\subsection{Reference-based}
The Reference-based baseline \citep{first_principle} extends the LOSS attack by calibrating the target model's loss with respect to a reference model trained on similar data but not necessarily the same data points. This helps to account for the intrinsic complexity of the target sample and reduces false negatives.

\begin{equation}
S(\mathbf{x}; {M}) = {Loss}(\mathbf{x}; {M})  - {Loss}(\mathbf{x}; {M}_\text{ref})
\end{equation}

\subsection{Zlib Entropy}
The Zlib Entropy baseline \citep{zlib} normalizes the target model's loss using the zlib compression size of the input sample. The idea is that the loss of member samples will have lower entropy and thus a smaller compression size compared to non-members.

\begin{equation}
S(\mathbf{x}; {M}) = \frac{{Loss}(\mathbf{x}; {M})}{\text{zlib}(\mathbf{x})}
\end{equation}

\subsection{Neighborhood Attack}
The Neighborhood Attack baseline \citep{mia_neighbor} estimates the curvature of the loss function around the target sample by comparing its loss to the average loss of its perturbed neighbors. The intuition is that member samples will have a lower loss compared to their neighbors, resulting in a larger difference.

\begin{equation}
S(\mathbf{x}; {M}) = {Loss}(\mathbf{x}; {M}) - \frac{1}{n}\sum_{i=1}^n {Loss}(\mathbf{\tilde{x}}_i; {M})
\end{equation}

\subsection{Min-K\%}
The Min-K\% baseline \citep{mia_mink} computes the membership score using the average log-likelihood of the $k\%$ of tokens with the lowest probabilities. This focuses on the least likely tokens, which are expected to have higher probabilities for member samples compared to non-members.

\begin{equation}
S(\mathbf{x}; {M}) = \frac{1}{\lvert \text{min-$k(\mathbf{x})$}\rvert} \sum_{x_i \in \text{min-$k(\mathbf{x})$}}-\log(p(x_i \mid x_1,...,x_{i-1}))
\end{equation}

\subsection{Min-K\%++}
The Min-K\%++ baseline \citep{mia_minkpp} is an extension of the Min-K\% that calibrates the next token log-likelihood with two factors: the mean ($\mu_{x_{<t}}$) and standard deviation ($\sigma_{x_{<t}}$) of the log-likelihood over all candidate tokens in the vocabulary. 

\begin{equation}
S_{\text{token}}(x_{<t},x_t; {M})=\frac{\log p(x_t|x_{<t};{M})-\mu_{x_{<t}}}{\sigma_{x_{<t}}},
\label{eq:minkpp_token}
\end{equation}
\begin{equation}
S(\mathbf{x}; {M})=\frac{1}{|\text{min-}k\%|}\sum_{(x_{<t},x_t)\in\text{min-}k\%}f_{\text{token}}(x_{<t},x_t; {M}).
\label{eq:minkpp}
\end{equation}
$\mu_{x_{<t}}$ and $\sigma_{x_{<t}}$ are the mean and standard deviation of the log-likelihoods over the model's vocabulary distribution given the prefix $x_{<t}$, respectively. The final membership score is obtained by averaging the normalized log-likelihoods of the $k\%$ of token sequences with the lowest scores (Equation \ref{eq:minkpp}).

\section{Additional Metrics Details}
\label{app:metric_detail}
\subsection{Area Under the ROC Curve (AUC)}
The area under the ROC curve (AUC) is a widely used metric for evaluating the performance of binary classification models, including MIAs. The ROC curve plots the true positive rate (TPR) against the false positive rate (FPR) at various decision thresholds. The TPR, also known as sensitivity or recall, is the proportion of actual positive samples (i.e., member samples) that are correctly identified as such. The FPR, on the other hand, is the proportion of actual negative samples (i.e., non-member samples) that are incorrectly identified as positive.

The AUC ranges from 0 to 1, with a value of 0.5 indicating a random classifier and a value of 1 indicating a perfect classifier. In the context of MIAs, a higher AUC value indicates that the attack is better at distinguishing between member and non-member samples across \textit{all possible} decision thresholds.  

\subsection{True Positive Rate at a Low False Positive Rate (TPR@low\%FPR)}
While the AUC provides an overall measure of an MIA's performance, it may not always be the most appropriate metric for practical applications. In many cases, the cost of false positives (i.e., incorrectly identifying a non-member sample as a member) can be much higher than the cost of false negatives (i.e., incorrectly identifying a member sample as a non-member). For example, in privacy-sensitive applications, falsely accusing an individual of being a member of a sensitive dataset can have severe consequences \citep{openai_vs_ny_time}.

To address this concern, we report the true positive rate at a low false positive rate (TPR@low\%FPR) \citep{first_principle}. In our experiments, we set the false positive rate threshold to 1\%, which means that we measure the proportion of member samples that are correctly identified as such while allowing only 1\% of non-member samples to be incorrectly identified as members. This metric provides a more stringent evaluation of an MIA's performance, focusing on its ability to correctly identify member samples while maintaining a low false positive rate.

\section{Additional Implementation Details}\label{app:implement_detail}
We use 16-bit floating-point precision for models larger than 60B to reduce computational requirement, and experiments are all conducted on 4 NVIDIA A6000 GPUs. In some experiments, we intentionally exceed the context window to test the limit, which might result in an out-of-memory (OOM) error. To ensure a fair evaluation, we also remove 12 data points from the member set for data balance, as this is a binary classification task. We report the best number of shot used for the main results in \Cref{app:best_shot} from the main results. It's worth noting that the Neighbor attack is significantly more computationally intensive than other methods \citep{mimir,mia_minkpp}, as it needs to iterate through the input's neighbor. Therefore, we obtain the Neighbor attack AUC results from \citet{mia_minkpp}. In contrast, \method is computationally efficient, as it only requires two LL calculations per sample, avoiding the need for expensive operations like building reference models \citep{first_principle,watson2021importance} or exploring neighboring samples \citep{mia_neighbor}.

\vspace{2em}
\begin{figure}[!h]
\centering
\section{Additional Model's Visualizations}\label{app:recall_visual_v1}
\includegraphics[width=\columnwidth]{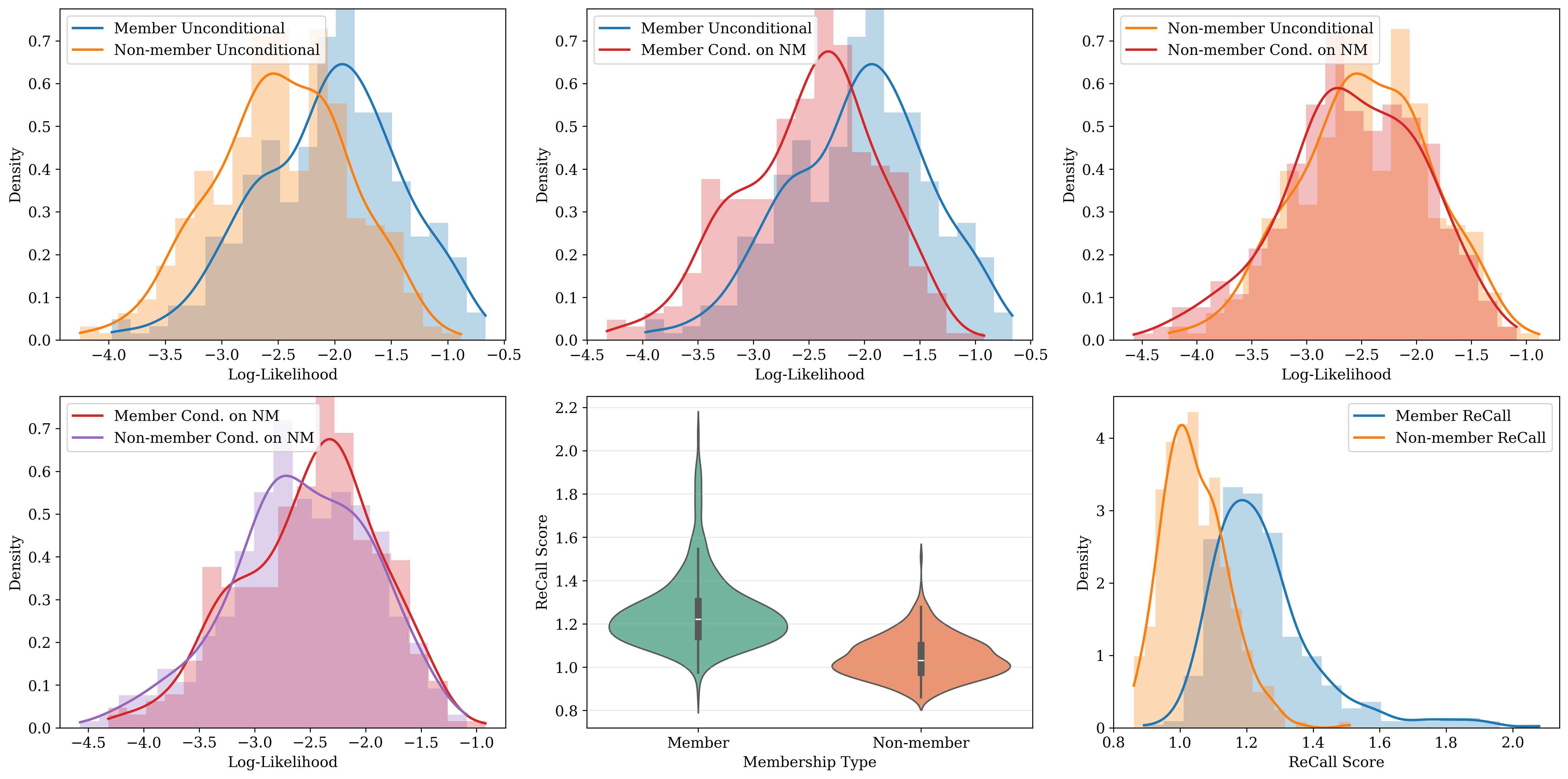}
\caption*{(a) NeoX-20B}
\vspace{1em}
\includegraphics[width=\columnwidth]{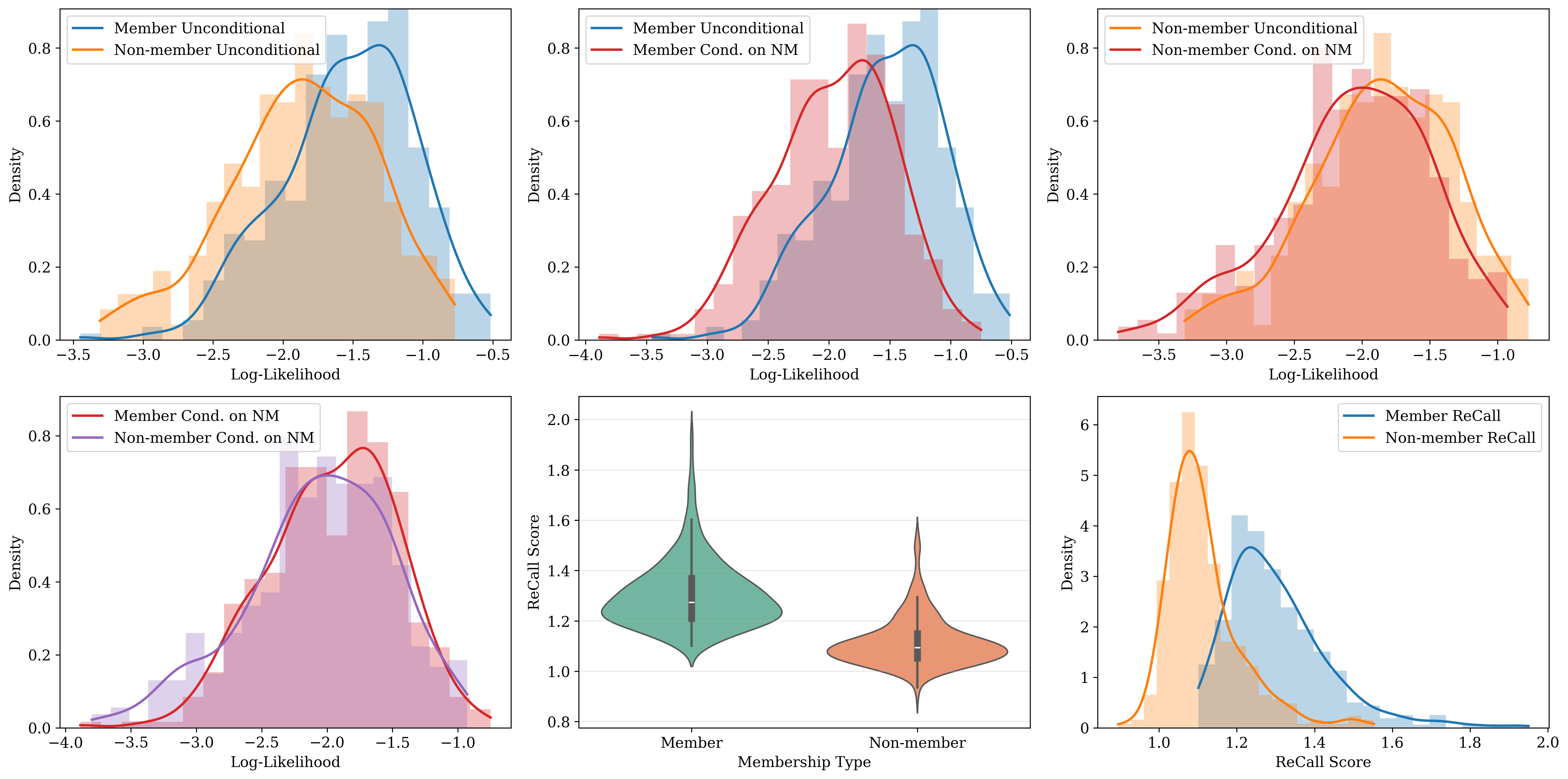}
\caption*{(b) LLaMA-13B}
\caption{Visualizations for \textbf{(a) NeoX-20B} and \textbf{(b) LLaMA-13B} for WIKIMIA with 5-shot results. Similar patterns are observed that members tend to be shifted further and have higher \method{} scores compared to non-members.}
\end{figure}
\newpage
\clearpage
\begin{table*}[ht!]
\begin{center} \scriptsize
\section{MIMIR 7-gram Results} \label{app:mimir_main_results_7}
\vspace{3em}
\subsection{AUC Results} 
\vspace{1em}
\vspace{1em}
\setlength{\tabcolsep}{0.7pt}
\begin{tabularx}{\textwidth}{l *{20}{>{\centering\arraybackslash}X}@{}}
    \toprule
    \multirow{2}{*}{}  & \multicolumn{5}{c}{\textbf{Wikipedia}} & \multicolumn{5}{c}{\textbf{Github}} & \multicolumn{5}{c}{\textbf{Pile CC}} & \multicolumn{5}{c}{\textbf{PubMed Central}} \\
    \cmidrule(lr){2-6}  \cmidrule(lr){7-11} \cmidrule(lr){12-16} \cmidrule(lr){17-21}
    \textbf{Method} & 160M & 1.4B & 2.8B & 6.9B & 12B
    & 160M & 1.4B & 2.8B & 6.9B & 12B
    & 160M & 1.4B & 2.8B & 6.9B & 12B
    & 160M & 1.4B & 2.8B & 6.9B & 12B
    \\
    \midrule
    
Loss &\textbf{62.6} & 65.7 & 66.4&68.0 & 69.0&84.1 &	87.2&88.0 & 88.8& 89.3&\textbf{53.1} &	\textbf{54.4}&54.8 &56.0 &56.4&\textbf{79.4} &	78.7&78.4 &78.5 & 78.4\\
Ref & 62.3 & \textbf{65.9}&\textbf{66.6} & \textbf{68.4}& 69.5&83.4 &	87.5& 88.3& 89.4&90.0 &52.9 &	\textbf{54.4}& \textbf{54.9}&56.2 & \textbf{56.7}& \textbf{79.4}&	78.8  &  78.4& 78.5 & 78.4 \\
Zlib & 57.3 &  62.0	&63.1 & 65.0&66.2 & \textbf{87.9}&\textbf{89.9}& \textbf{90.6}&\textbf{91.3} & \textbf{91.7}& 51.4&	53.2&53.7 & 54.8& 55.2& 78.0&	77.6  & 77.3 & 77.5 & 77.4 \\
Min-K\% & 60.8 & 64.8& 65.9& 68.0&69.6 &82.8 &	87.0& 87.9& 88.8& 89.4& 52.6&	54.0&54.6 &56.0 & 56.2& 77.9&	78.6  & 78.2 & 78.8 & 78.9 \\
Min-K\%++ & 62.3 & 63.6 & 64.7 & 68.1 & \textbf{70.0} & 83.1 & 83.2 & 84.8 & 85.5 & 86.9 & 51.5 & 52.6 & 53.7 & 55.7 & 56.1 & 76.9 & 66.1 & 66.6 & 68.3 & 68.8 \\
\rowcolor{gray!15} \method &61.3 &64.9	& 66.0& 67.5& 69.0&83.5 &87.6	& 88.6&90.7 & 91.6&51.8 &53.6&54.7 &\textbf{56.5} & \textbf{56.7}& 76.4&	\textbf{78.9}  & \textbf{78.9}& \textbf{81.3}&\textbf{79.8}\\
    
    \toprule
    \multirow{2}{*}{}  & \multicolumn{5}{c}{\textbf{ArXiv}} & \multicolumn{5}{c}{\textbf{DM Mathematics}} & \multicolumn{5}{c}{\textbf{HackerNews}} & \multicolumn{5}{c}{\textbf{Average}}\\
    \cmidrule(lr){2-6}  \cmidrule(lr){7-11} \cmidrule(lr){12-16} \cmidrule(lr){17-21}
    \textbf{Method} & 160M & 1.4B & 2.8B & 6.9B & 12B
    & 160M & 1.4B & 2.8B & 6.9B & 12B
    & 160M & 1.4B & 2.8B & 6.9B & 12B
    & 160M & 1.4B & 2.8B & 6.9B & 12B
    \\
    \midrule

% Loss & 75.5 & 77.5	&78.0 &79.0 &79.4 & 93.6&91.7	&91.4 &91.5 &91.4 &58.6 &59.6 &60.5  & 61.2 &62.0  \\
% Ref & 75.5 &  77.8& 78.2& 79.3&79.8 & 93.7&	91.4	& 91.0&91.1& 90.9&58.6 &59.7& 60.6 & 61.3 &62.2  \\
% Zlib & 74.9 &  76.9	&77.3 &78.1 &78.5 & 81.8&81.2	& 81.6& 81.4&81.3 & 57.8&58.9  &59.5  & 59.9 &  60.6\\
% % Neighbor &  &  &  &  & / &  &  &  &  & / &  &  &  &  & / &  &  &  &  & / \\
% Min-K\% & 70.6 &  74.2		& 75.3&76.7 &77.7 &92.9 &92.5	& 92.4&92.4 &92.2 &55.5 &	56.8  &58.0  &59.0 &60.4  \\
% Min-K\%++ & 50.0 & 62.4	&64.6 & 67.0&69.0 & 50.0&67.0	& 69.1& 64.3& 66.3&50.0 &55.8  & 57.5 &58.8  & 60.4 &  \\

% \rowcolor{gray!15} \method  &75.4& 76.6&77.8 & 77.0&77.6 & 95.3&	94.3	& 93.9&92.9 &92.1 &59.4 &60.1&60.8 &63.1&63.3 \\

Loss & \textbf{75.5}& 77.5 & 78.0 & 79.0 & 79.4 & 93.6 & 91.7 & 91.4 & 91.5 & 91.4 & 58.6 & 59.6 & 60.5 & 61.2 & 62.0 & \textbf{72.4} & 73.5 & 73.9 & 74.7 & 75.1 \\
Ref & \textbf{75.5} & \textbf{77.8} & \textbf{78.2} & \textbf{79.3} & \textbf{79.8} & 93.7 & 91.4 & 91.0 & 91.1 & 90.9 & 58.6 & 59.7 & 60.6 & 61.3 & 62.2 & 72.3 & 73.6 & 74.0 & 74.9 & 75.4  \\
Zlib & 74.9 & 76.9 & 77.3 & 78.1 & 78.5 & 81.8 & 81.2 & 81.6 & 81.4 & 81.3 & 57.8 & 58.9 & 59.5 & 59.9 & 60.6 & 69.9 & 71.4 & 71.9 & 72.6 & 73.0  \\
Min-K\% & 70.6 & 74.2 & 75.3 & 76.7 & 77.7 & 92.9 & 92.5 & 92.4 & 92.4 & \textbf{92.2} & 55.5 & 56.8 & 58.0 & 59.0 & 60.4 & 70.4 & 72.6 & 73.2 & 74.2 & 74.9 \\
Min-K\%++ & 70.0 & 62.4 & 64.6 & 67.0 & 69.0 & 90.9 & 67.0 & 69.1 & 64.3 & 66.3 & 58.8 & 55.8 & 57.5 & 58.8 & 60.4 & 70.5 & 64.4 & 65.9 & 66.8 & 68.2 \\

\rowcolor{gray!15}\method{} & 75.4 & 76.6 & 77.8 & 77.0 & 77.6 & \textbf{95.3} & \textbf{94.3} & \textbf{93.9} & \textbf{92.9} & 92.1 & \textbf{59.4} & \textbf{60.1} & \textbf{60.8} & \textbf{63.1} & \textbf{63.3} & 71.9 & \textbf{73.7}  & \textbf{74.4} & \textbf{75.6} & \textbf{75.7} \\

\bottomrule
\end{tabularx}
\caption{AUC results on the challenging MIMIR benchmark \cite{mimir} in the 7-gram setting. The best result across all methods is \textbf{bolded} in each column. \method{} outperforms all baselines on the 1.4B, 2.8B, 6.9B, and 12B models in average, demonstrating its effectiveness in detecting pretraining data even when the distribution shift between members and non-members is minimized.}

\end{center}
\end{table*}

\begin{table*}[ht!]
\begin{center} \scriptsize
\subsection{TPR@1\%FPR 
 Result}\label{app:mimir_tpr_7}
\vspace{1em}

\setlength{\tabcolsep}{0.7pt}
\begin{tabularx}{\textwidth}{l *{20}{>{\centering\arraybackslash}X}@{}}
    \toprule
    \multirow{2}{*}{}  & \multicolumn{5}{c}{\textbf{Wikipedia}} & \multicolumn{5}{c}{\textbf{Github}} & \multicolumn{5}{c}{\textbf{Pile CC}} & \multicolumn{5}{c}{\textbf{PubMed Central}} \\
    \cmidrule(lr){2-6}  \cmidrule(lr){7-11} \cmidrule(lr){12-16} \cmidrule(lr){17-21}
    \textbf{Method} & 160M & 1.4B & 2.8B & 6.9B & 12B
    & 160M & 1.4B & 2.8B & 6.9B & 12B
    & 160M & 1.4B & 2.8B & 6.9B & 12B
    & 160M & 1.4B & 2.8B & 6.9B & 12B
    \\
    \midrule
    
Loss & \textbf{6.9} & 11.0 & 11.5 & 13.4 & 13.2 & 27.0 & 45.3 & 47.7 & 51.2 & 50.8 & 2.4 & 3.9 & 4.6 & 5.4 & 5.7 & 16.1 & 16.9 & \textbf{19.6} & 13.6 & 14.4  \\
Ref & 6.2 & \textbf{12.4} & \textbf{12.1} & \textbf{14.3} & 13.1 & 20.3 & 43.8 & 52.0 & 55.9 & 55.5 & 1.9 & 3.9 & 4.5 & 5.6 & 6.3 & 15.0 & 15.0 & 17.7 & 12.3 & 14.2   \\
Zlib & 3.8 & 7.8 & 8.9 & 10.1 & 10.6 & \textbf{56.2} & \textbf{57.4} & \textbf{62.1} & \textbf{61.7} & \textbf{58.6} & \textbf{2.6} & \textbf{4.6} & \textbf{5.5} & \textbf{6.7} & \textbf{7.8} & 16.9 & 14.4 & 14.8 & 12.7 & 10.0  \\
Min-K\% & 7.2 & 10.3 & 11.8 & 13.9 & 12.6 & 30.9 & 46.9 & 50.0 & 52.3 & 52.7 & 2.4 & 4.5 & 5.0 & 5.4 & 5.9 & \textbf{19.6} & \textbf{19.2} & 19.2 & 21.5 & \textbf{22.8}  \\
Min-K\%++ & 5.5 & 7.1 & 10.0 & 11.0 & 13.2 & 29.6 & 30.5 & 32.4 & 39.5 & 38.7 & 2.1 & 3.1 & 3.2 & 4.5 & 4.5 & 15.0 & 9.8 & 10.2 & 8.6 & 14.0  \\
\rowcolor{gray!15} \method{} & 6.7 & 10.7 & 11.7 & 13.6 & \textbf{15.7} & 32.3 & 48.8 & 49.6 & 52.7 & 56.2 & 2.1 & 2.9 & 4.0 & 4.9 & 4.8 & 16.3 & 13.8 & 19.4 & \textbf{24.4} & 22.5  \\
    
    \toprule
    \multirow{2}{*}{}  & \multicolumn{5}{c}{\textbf{ArXiv}} & \multicolumn{5}{c}{\textbf{DM Mathematics}} & \multicolumn{5}{c}{\textbf{HackerNews}} & \multicolumn{5}{c}{\textbf{Average}}\\
    \cmidrule(lr){2-6}  \cmidrule(lr){7-11} \cmidrule(lr){12-16} \cmidrule(lr){17-21}
    \textbf{Method} & 160M & 1.4B & 2.8B & 6.9B & 12B
    & 160M & 1.4B & 2.8B & 6.9B & 12B
    & 160M & 1.4B & 2.8B & 6.9B & 12B
    & 160M & 1.4B & 2.8B & 6.9B & 12B
    \\
    \midrule

Loss & 8.6 & 11.3 & 16.0 & 16.6 & 17.0 & 67.5 & 29.9 & 14.3 & 14.3 & 14.3 & 2.2 & 1.9 & 1.9 & 2.8 & 1.9 & 18.7 & 17.2 & 16.5 & 16.8 & 16.8  \\
Ref & 7.4 & 12.3 & 17.8 & 17.8 & 18.4 & 71.4 & 26.0 & 6.5 & 5.2 & 3.9 & 2.4 & 1.9 & 1.9 & 2.8 & 2.1 & 17.8 & 16.5 & 16.1 & 16.3 & 16.2   \\
Zlib & 4.7 & 9.0 & 12.3 & 15.4 & 16.2 & 19.5 & 14.3 & 7.8 & 6.5 & 6.5 & \textbf{2.8} & \textbf{2.4} & \textbf{2.7} & 2.8 & 3.3 & 15.2 & 15.7 & 16.3 & 16.6 & 16.1   \\
Min-K\% & 9.4 & 14.3 & \textbf{21.3} & \textbf{21.1} & \textbf{21.3} & 66.2 & \textbf{61.0} & \textbf{44.2} & \textbf{39.0} & \textbf{37.7} & 1.9 & 1.7 & 1.1 & 2.4 & 1.9 & 19.7 & \textbf{22.6} & \textbf{21.8} & \textbf{22.2} & \textbf{22.1}  \\
Min-K\%++ & \textbf{10.0} & 3.1 & 4.9 & 5.5 & 6.1 & 20.3 & 16.9 & 19.5 & 10.4 & 15.6 & 1.8 & 1.9 & 1.4 & 2.4 & 2.4 & 12.0 & 10.3 & 11.7 & 11.7 & 13.5  \\
\rowcolor{gray!15}\method{} & 5.7 & \textbf{15.6} & 15.6 & 17.4 & 14.1 &\textbf{79.2} & 44.2 & 31.2 & 22.1 & 15.6 & 1.1 & 1.7 & \textbf{2.7} & \textbf{4.9} & \textbf{4.6} & \textbf{20.5} & 19.7 & 19.2 & 20.0 & 19.1  \\
    
\bottomrule
\end{tabularx}
\caption{TPR@1\%FPR  results on the challenging MIMIR benchmark \cite{mimir} in 7-gram setting. The best result across all methods is \textbf{bolded} in each column.
}

\end{center}
\end{table*}

\section{Synthetic Prefixes Generation
}\label{app:gpt_prompt}
\begin{tcolorbox}[title=GPT-4o Prompt Template]\small
Generate a passage that is similar to the given text in length, domain, and style.
\\ \\
Given text: \{\textit{a data point (could be member or non-member)}\} 
\\ \\
New passage:

\end{tcolorbox}

% \begin{tcolorbox}[title=Setting 2]
% \small
% Generate 12 passages that are similar to the given text in length, domain, and style.
% \\ \\
% Given text: \{\textit{target data point}\} 
% \\ \\
% New passages:

% \end{tcolorbox}

\begin{table*}[ht!]
\centering
\section{WikiMIA TPR@1\%FPR 
 Results}\label{app:wikimia_tpr}
\vspace{1em}
\resizebox{0.95\textwidth}{!}{%
\begin{tabular}{llcccccccc}
\toprule
\textbf{Len.} & \textbf{Method} & \textbf{Mamba-1.4B} & \textbf{Pythia-6.9B} & \textbf{LLaMA-13B} & \textbf{NeoX-20B} & \textbf{LLaMA-30B} & \textbf{OPT-66B} & \textbf{Average} \\
\midrule
\multirow{6}{*}{32}
& Loss & 4.5 & 6.1 & 4.8 &10.4  &4.3  & 6.4 & 6.1\\ 
 & Ref & 4.5 & 6.9 & 5.9& 10.1 &  2.7& 6.7 & 6.1\\ 
 & Zlib & 4.0 & 4.8 &5.6 & 9.1 & 4.8 &5.6 & 5.7\\ 
 % & Neighbor &  &  &  &  &  &  &  \\ 
 & Min-K\% & 6.7 & 8.8 &5.1  &10.7  &4.5  & \textbf{9.1} & 7.5\\ 
 & Min-K\%++ &4.3  & 5.9 & 10.4 & 6.1 & 9.3 & 3.7 & 6.6\\ 

% \rowcolor{gray!15}\cellcolor{white}& \method (average) &  &  &  &  &  &  & \\
\rowcolor{gray!15}\cellcolor{white}& \method & \textbf{ 11.2} &\textbf{  28.5} & \textbf{ 13.3} &\textbf{ 25.3 } & \textbf{ 18.4} & 8.3 & \textbf{ 17.5}\\
\midrule

\multirow{6}{*}{64} 
& Loss & 3.3 &3.3  & 4.9 & 4.5 & 6.1 &4.1  & 4.4\\ 
 & Ref & 2.8 & 3.3 & 4.1 & 4.9 & 6.5 &4.5  &  4.4\\ 
 & Zlib & 6.1 & 6.9 & 8.9 & 7.7 &  10.6& 9.8 & 8.3\\ 
 % & Neighbor &  &  &  &  &  &  & \\ 
 & Min-K\% & 6.9 &6.5  &6.5  & 5.7 & 8.1 &10.2  & 7.3\\ 
 & Min-K\%++ & 7.3 & 11.8&15.4  & \textbf{10.2}  & 6.9 &  \textbf{11.8}& 10.6\\ 

% \rowcolor{gray!15}\cellcolor{white}& \method (average) &  &  &  &  &  &  &  \\
\rowcolor{gray!15}\cellcolor{white}& \method  & \textbf{ 11.0 }& \textbf{ 20.7 }&\textbf{ 30.1}  & 6.9 &  \textbf{ 18.3}& 5.3 &  \textbf{ 15.4}\\
\midrule
\multirow{6}{*}{128} 
& Loss & 1.0 & 3.0 & 7.1 & 4.0 & 1.0 & 4.0 & 3.4\\ 
 & Ref & 1.0 & 3.0 & 8.1 &4.0  & 0.0 &4.0  & 3.4\\ 
 & Zlib & \textbf{ 6.1} &6.1  & 10.1 & 5.1 & \textbf{ 2.0 }&\textbf{9.1}  & 6.4\\ 
 % & Neighbor &  &  &  &  &  &  & \\ 
 & Min-K\% & 3.0 & 4.0 & 8.1 & 3.0 & \textbf{ 2.0} & 4.0 & 4.0\\ 
 & Min-K\%++ &2.0  &8.1  & 8.1 &1.0  & 0.0 & 0.0 & 3.2\\ 
% \rowcolor{gray!15}\cellcolor{white}& \method (average) &  &  &  &  &  &  &  \\
\rowcolor{gray!15}\cellcolor{white}& \method  &4.0  & \textbf{33.3} & \textbf{26.3}  & \textbf{30.3} &  1.0& 6.1 & \textbf{16.9} \\
\bottomrule
\end{tabular}
}
\caption{
TPR@1\%FPR results on WikiMIA benchmark. \textbf{Bolded} numbers show the best result within each column. Overall, \method{} consistently achieves the highest average TPR@1\%FPR scores across all input lengths, demonstrating its effectiveness in detecting pretraining data with high precision.}
\end{table*}

\begin{table*}[ht!]
\begin{center} \scriptsize
\section{MIMIR 13-gram TPR@1\%FPR 
 Result}\label{app:mimir_tpr_13}
\vspace{1em}

\setlength{\tabcolsep}{0.7pt}
\begin{tabularx}{\textwidth}{l *{20}{>{\centering\arraybackslash}X}@{}}
    \toprule
    \multirow{2}{*}{}  & \multicolumn{5}{c}{\textbf{Wikipedia}} & \multicolumn{5}{c}{\textbf{Github}} & \multicolumn{5}{c}{\textbf{Pile CC}} & \multicolumn{5}{c}{\textbf{PubMed Central}} \\
    \cmidrule(lr){2-6}  \cmidrule(lr){7-11} \cmidrule(lr){12-16} \cmidrule(lr){17-21}
    \textbf{Method} & 160M & 1.4B & 2.8B & 6.9B & 12B
    & 160M & 1.4B & 2.8B & 6.9B & 12B
    & 160M & 1.4B & 2.8B & 6.9B & 12B
    & 160M & 1.4B & 2.8B & 6.9B & 12B
    \\
    \midrule
    
Loss & 0.7 & 0.8 & 0.6 & 0.7 & 0.9 & 16.0 & 19.7 & 22.2 & 22.5 & 23.1 & 0.4 & 0.5 & 0.8 & 0.8 & 0.8 & \textbf{0.8} & 0.8 & 0.8 & 0.7 & 0.4  \\
Ref & 1.1 & 0.5 & \textbf{0.7} & 0.9 & 0.9 & 17.1 & 9.2 & 10.0 & 11.6 & 13.1 & 0.6 & \textbf{0.6} & 0.7 & 0.9 & 0.9 & \textbf{0.8} & \textbf{0.9} & 0.6 & 0.6 & 0.4   \\
Zlib & 0.8 & 0.7 & \textbf{0.7} & 0.9 & \textbf{1.0} & \textbf{17.4} & \textbf{23.0} & \textbf{24.0} & \textbf{26.0} & \textbf{25.9} & 0.5 & \textbf{0.6} & 0.9 & 1.1 & 1.1 & 0.5 & 0.5 & 0.3 & 0.6 & 0.5  \\
Min-K\% & 1.1 & 0.8 & 0.6 & 0.7 & 0.9 & 15.2 & 20.3 & 21.6 & 22.7 & 23.2 & 0.4 & 0.5 & 0.7 & 0.7 & 0.9 & 0.7 & 0.4 & 0.6 & 0.6 & 0.7  \\
Min-K\%++ & 0.9 & 0.7 & 0.6 & \textbf{1.1} & \textbf{1.0} & 13.4 & 18.2 & 18.8 & 21.5 & 23.6 & \textbf{0.7} & \textbf{0.6} & \textbf{1.1} & \textbf{1.2} & \textbf{1.4} & 0.6 & 0.6 & \textbf{1.0} & \textbf{1.1} & \textbf{1.2}  \\
\rowcolor{gray!15} \method{} & \textbf{1.3} & \textbf{0.9} & \textbf{0.7} & 0.7 & 0.8 & 11.6 & 21.5 & 23.1 & 22.5 & 24.6 & \textbf{0.7} & 0.4 & 0.5 & 0.9 & 1.1 & 0.4 & 0.7 & 0.4 & 0.3 & 0.5  \\
    
    \toprule
    \multirow{2}{*}{}  & \multicolumn{5}{c}{\textbf{ArXiv}} & \multicolumn{5}{c}{\textbf{DM Mathematics}} & \multicolumn{5}{c}{\textbf{HackerNews}} & \multicolumn{5}{c}{\textbf{Average}}\\
    \cmidrule(lr){2-6}  \cmidrule(lr){7-11} \cmidrule(lr){12-16} \cmidrule(lr){17-21}
    \textbf{Method} & 160M & 1.4B & 2.8B & 6.9B & 12B
    & 160M & 1.4B & 2.8B & 6.9B & 12B
    & 160M & 1.4B & 2.8B & 6.9B & 12B
    & 160M & 1.4B & 2.8B & 6.9B & 12B
    \\
    \midrule

Loss & 0.5 & 0.3 & 0.6 & 0.7 & 0.7 & 0.7 & 0.6 & 1.0 & \textbf{1.1} & \textbf{1.1} & 0.8 & 0.6 & 0.6 & 0.7 & 0.8 & 2.8 & 3.3 & 3.8 & 3.9 & 4.0  \\
Ref & 0.6 & 0.3 & 0.7 & 0.8 & 1.0 & 0.7 & \textbf{1.0} & \textbf{1.3} & 1.0 & 1.0 & \textbf{1.1} & 0.6 & 0.6 & 0.7 & 1.0 & \textbf{3.1} & 1.9 & 2.1 & 2.4 & 2.6   \\
Zlib & 0.5 & 0.3 & 0.4 & 0.4 & 0.7 & \textbf{1.1} & 0.7 & 0.9 & 0.9 & 0.9 & 1.0 & 0.9 & 1.3 & \textbf{1.3} & 1.1 & \textbf{3.1} & \textbf{3.8} & \textbf{4.1} & \textbf{4.5} & \textbf{4.5}   \\
Min-K\% & 0.5 & 0.2 & 0.5 & 0.4 & 0.8 & 0.7 & 0.5 & 0.2 & 0.4 & 0.4 & 0.7 & 0.8 & 0.7 & 0.9 & 0.9 & 2.8 & 3.4 & 3.6 & 3.8 & 4.0  \\
Min-K\%++ & 0.5 & \textbf{1.3} & \textbf{1.4} & 1.0 & 1.8 & 0.6 & \textbf{1.0} & 1.2 & 0.4 & 0.9 & 0.7 & 0.4 & 1.0 & \textbf{1.3} & 0.5 & 2.5 & 3.3 & 3.6 & 3.9 & 4.3  \\
\rowcolor{gray!15}\method{} & \textbf{1.1} & 0.8 & 0.9 & \textbf{1.4} & \textbf{2.4} & 0.3 & 0.9 & 0.6 & 0.3 & 0.0 & 1.0 & \textbf{1.6} & \textbf{1.7} & 1.1 & \textbf{1.5} & 2.3 & \textbf{3.8} & 4.0 & 3.9 & 4.4  \\
   
\bottomrule
\end{tabularx}
\caption{TPR@1\%FPR  results on the challenging MIMIR benchmark \cite{mimir} in 13-gram setting. The best result across all methods is \textbf{bolded} in each column.
}

\end{center}
\end{table*}

\begin{table*}[ht!]
\centering
\small
\section{Additional Synthetic Prefix Results}\label{app:gpt_prefix}
\vspace{1em}

\begin{tabular}{lccc}
\toprule
\textbf{Model} & \textbf{Len. 32} & \textbf{Len. 64} & \textbf{Len. 128} \\
\midrule
Pythia-6.9B - Synthetic  & 83.7 & 87.1 & 83.0 \\
Pythia-6.9B - Real  & 91.6 & 93.0 & 92.6 \\
\midrule
Pythia-12B - Synthetic  & 85.4 & 90.3 & 86.4 \\
Pythia-12B - Real  & 88.2 & 88.8 & 87.8 \\
\midrule
LLaMA-13B - Synthetic  & 89.2 & 93.2 & 90.5 \\
LLaMA-13B - Real  & 92.2 & 95.2 & 92.5 \\
\bottomrule
\end{tabular}
\caption{The performance of synthetic and real prefixes across different models (Pythia-6.9B, Pythia-12B, and LLaMA-13B). \method{} achieves comparable performance even with synthetic prefixes generated by GPT-4.}
\end{table*}

\begin{figure}[ht!]
\centering
\section{Additional Prefix Visualizations}\label{app:member_prefix_visual}
\vspace{1em}

\includegraphics[width=\textwidth]{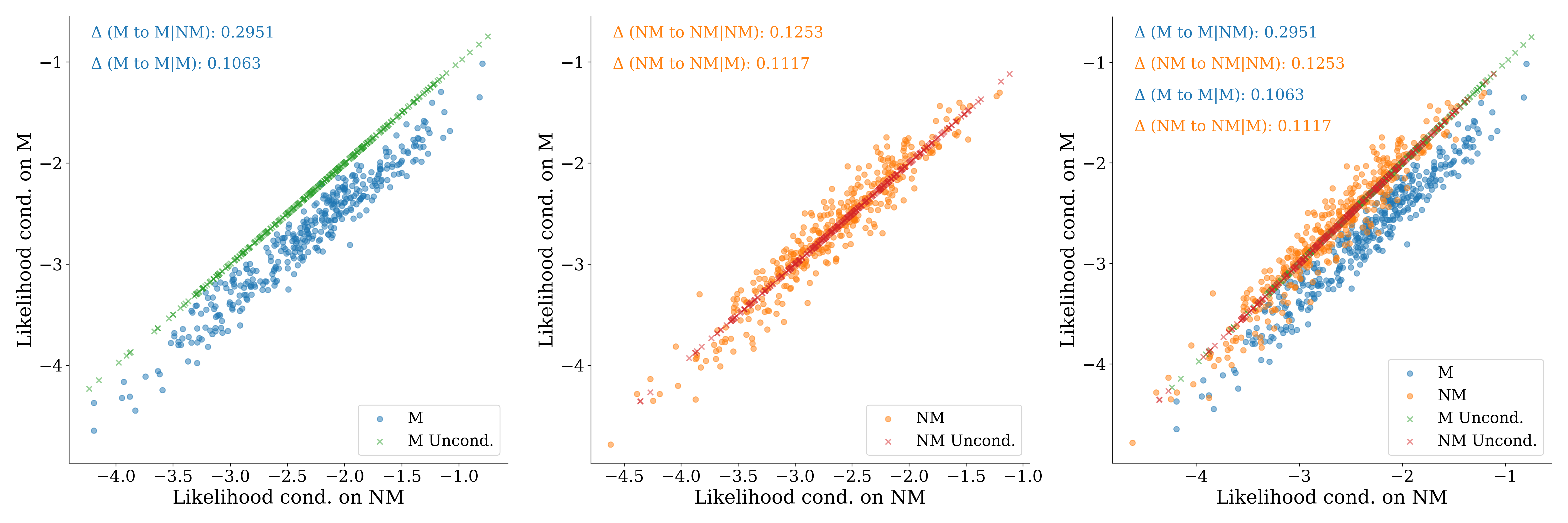}
\caption{Conditional LL for members and non-members with member and non-member prefix comparison. Their unconditional LL are in the diagonal line. Conditioning both member and non-member data with member prefix do not yield significant changes in LL.}
\label{fig:fig_1_v1}
\end{figure}

\begin{figure}[ht!]
\centering
\section{Additional Token-level Results}\label{app:token_level_visual}
\vspace{1em}
\includegraphics[width=\columnwidth]{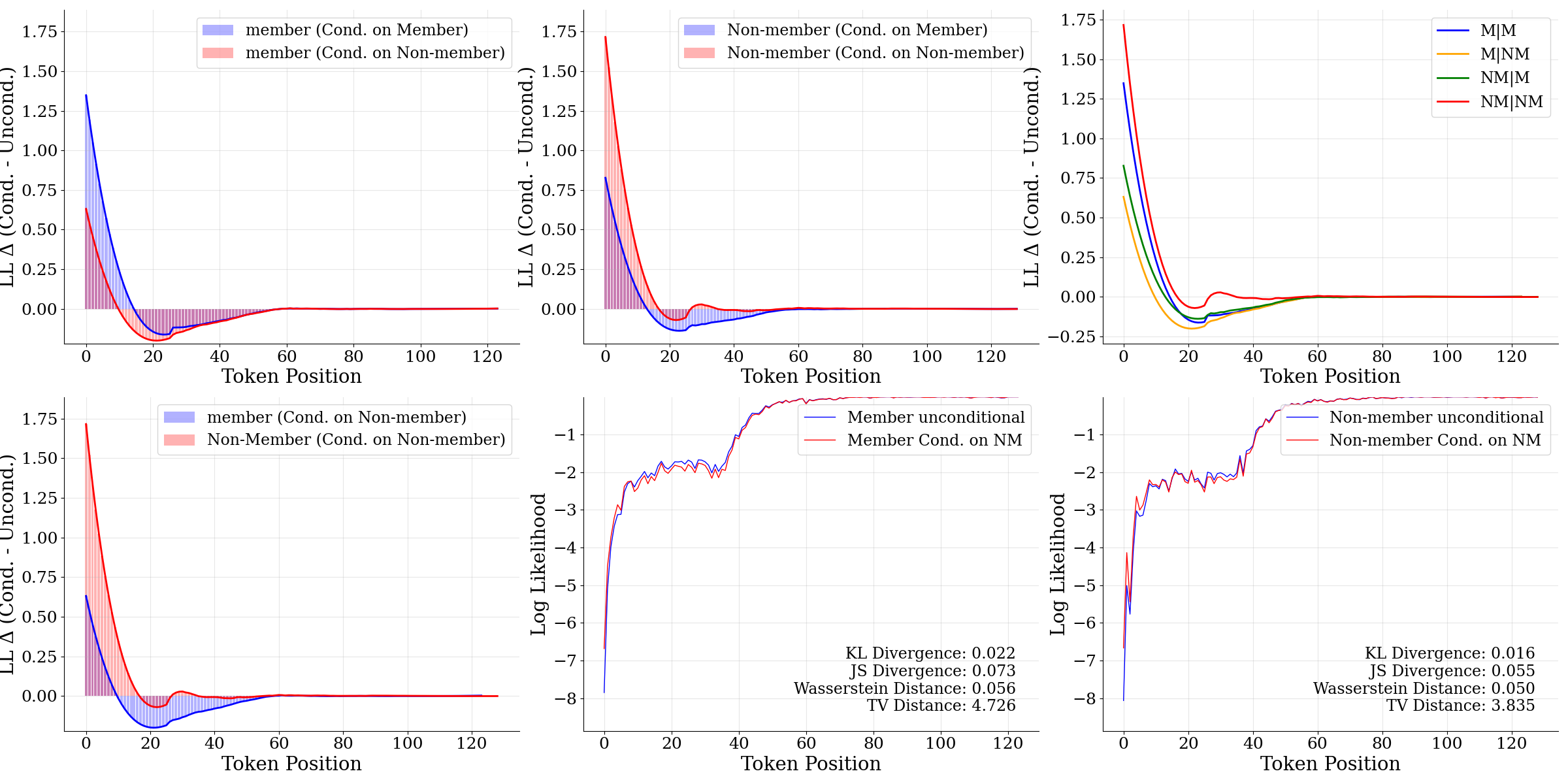}
\caption{Average token-level log-likelihood changes for member (M) and non-member (NM) data points when prefixed with member and non-member context. The largest changes occur in the beginning tokens, and data points experience the most dominant changes when prefixed with context from the same membership category. Member and non-member data exhibit the largest differences when prefixed with non-member context, consistent with the findings in \Cref{fig:token_levelL_visual}.}
\end{figure}
\begin{table*}[ht!]
\centering
\section{Best Number of Shot}\label{app:best_shot}
\vspace{1em}
\subsection{WikiMIA}
\resizebox{0.95\textwidth}{!}{%
\begin{tabular}{llccccccc}
\toprule
\textbf{Len.}  & \textbf{Mamba-1.4B} & \textbf{Pythia-6.9B} & \textbf{LLaMA-13B} & \textbf{NeoX-20B} & \textbf{LLaMA-30B} & \textbf{OPT-66B} & \textbf{Average} \\
\midrule
\multirow{1}{*}{32}
 &7  &  7& 6 &  7&  6& 6 & 6.5 & \\ 
\multirow{1}{*}{64} 
 & 10 & 9 & 8 & 12 & 12 &9  & 10 & \\ 
\multirow{1}{*}{128} 
 & 11 & 7 & 6 & 6 & 9 & 4 & 7.2 & \\ 
\bottomrule
\end{tabular}
}
\caption{Report on the best number of shot is used in WikiMIA main result. }
\end{table*}

\vspace*{-2cm} % Adjust this value as needed
\begin{table*}[t!]
\centering
\scriptsize
\subsection{MIMIR 13-gram}
\vspace{1em}
\setlength{\tabcolsep}{0.7pt}
\begin{tabularx}{\textwidth}{l *{20}{>{\centering\arraybackslash}X}@{}}
    \toprule
    \multirow{2}{*}{}  & \multicolumn{5}{c}{\textbf{Wikipedia}} & \multicolumn{5}{c}{\textbf{Github}} & \multicolumn{5}{c}{\textbf{Pile CC}} & \multicolumn{5}{c}{\textbf{PubMed Central}} \\
    \cmidrule(lr){2-6}  \cmidrule(lr){7-11} \cmidrule(lr){12-16} \cmidrule(lr){17-21}
    \textbf{Method} & 160M & 1.4B & 2.8B & 6.9B & 12B
    & 160M & 1.4B & 2.8B & 6.9B & 12B
    & 160M & 1.4B & 2.8B & 6.9B & 12B
    & 160M & 1.4B & 2.8B & 6.9B & 12B
    \\
    \midrule
    
 \method & 12 & 8 & 1 & 8 & 8 & 10 & 3 & 3 & 7 & 7 & 9 & 9 & 12 & 5 & 11 & 12 & 7 & 11 & 1 & 1  \\
    
    \toprule
    \multirow{2}{*}{}  & \multicolumn{5}{c}{\textbf{ArXiv}} & \multicolumn{5}{c}{\textbf{DM Mathematics}} & \multicolumn{5}{c}{\textbf{HackerNews}} & \multicolumn{5}{c}{\textbf{Average}}\\
    \cmidrule(lr){2-6}  \cmidrule(lr){7-11} \cmidrule(lr){12-16} \cmidrule(lr){17-21}
    \textbf{Method} & 160M & 1.4B & 2.8B & 6.9B & 12B
    & 160M & 1.4B & 2.8B & 6.9B & 12B
    & 160M & 1.4B & 2.8B & 6.9B & 12B
    & 160M & 1.4B & 2.8B & 6.9B & 12B
    \\
    \midrule
    
 \method & 5 & 6 & 4 & 6 & 6 & 1 & 1 & 1 & 1 & 1 & 5 & 3 & 3 & 5 & 7 & 7.7 & 5.3 & 5.0 & 4.7 & 5.9  \\
    
\bottomrule
\end{tabularx}
\caption{Report on the best number of shot used in MIMIR 13-gram main result.}
\label{tab:mimir_best_shot_13}
\end{table*}

\begin{table*}[t!]
\centering
\scriptsize
\subsection{MIMIR 7-gram}
\vspace{1em}
\setlength{\tabcolsep}{0.7pt}
\begin{tabularx}{\textwidth}{l *{20}{>{\centering\arraybackslash}X}@{}}
    \toprule
    \multirow{2}{*}{}  & \multicolumn{5}{c}{\textbf{Wikipedia}} & \multicolumn{5}{c}{\textbf{Github}} & \multicolumn{5}{c}{\textbf{Pile CC}} & \multicolumn{5}{c}{\textbf{PubMed Central}} \\
    \cmidrule(lr){2-6}  \cmidrule(lr){7-11} \cmidrule(lr){12-16} \cmidrule(lr){17-21}
    \textbf{Method} & 160M & 1.4B & 2.8B & 6.9B & 12B
    & 160M & 1.4B & 2.8B & 6.9B & 12B
    & 160M & 1.4B & 2.8B & 6.9B & 12B
    & 160M & 1.4B & 2.8B & 6.9B & 12B
    \\
    \midrule
    
 \method & 1 & 2 & 10 & 9 & 9 & 11 & 9 & 7 & 8 & 8 & 12 & 4 & 4 & 4 & 4 & 12 & 7 & 4 & 6 & 1  \\
    
    \toprule
    \multirow{2}{*}{}  & \multicolumn{5}{c}{\textbf{ArXiv}} & \multicolumn{5}{c}{\textbf{DM Mathematics}} & \multicolumn{5}{c}{\textbf{HackerNews}} & \multicolumn{5}{c}{\textbf{Average}}\\
    \cmidrule(lr){2-6}  \cmidrule(lr){7-11} \cmidrule(lr){12-16} \cmidrule(lr){17-21}
    \textbf{Method} & 160M & 1.4B & 2.8B & 6.9B & 12B
    & 160M & 1.4B & 2.8B & 6.9B & 12B
    & 160M & 1.4B & 2.8B & 6.9B & 12B
    & 160M & 1.4B & 2.8B & 6.9B & 12B
    \\
    \midrule
\method & 12 & 11 & 12 & 11 & 11 & 12 & 12 & 11 & 10 & 10 & 11 & 10 & 8 & 8 & 8 & 10.1 & 7.9 & 8.0 & 8.0 & 7.3  \\
    
\bottomrule
\end{tabularx}
\caption{Report on the best number of shot used in MIMIR 7-gram main result.}
\label{tab:mimir_best_shot_7}
\end{table*}
% \section{Additional Member Prefix Results}\label{app:member_prefix_results}

\end{document}